\documentclass{article}

\usepackage{arxiv}
\usepackage[utf8]{inputenc} 
\usepackage[T1]{fontenc}    
\usepackage{hyperref}       
\usepackage{url}            
\usepackage{booktabs}       
\usepackage{amsfonts}       
\usepackage{nicefrac}       
\usepackage{microtype}      
\usepackage{lipsum}		
\usepackage{doi}

\usepackage{bm}
\usepackage{graphicx}
\usepackage{amsmath, amssymb,algorithm}
\usepackage{comment}
\usepackage{tabularx}
\usepackage{cuted}
\usepackage{xcolor}
\usepackage{multicol}

\DeclareMathOperator*{\argmin}{argmin}

\newcommand{\vect}[1]{\boldsymbol{#1}}

\newdimen\ocf
\newdimen\tcf
\title{On Learning what to Learn: \\
heterogeneous observations of dynamics and establishing (possibly causal) relations among them}

\author{ \hspace{1mm}{David W. Sroczynski} \\
	Department of Chemical and Biological Engineering\\
	Princeton University\\
	Princeton NJ, USA \\
    \And
    \hspace{1mm}{Felix Dietrich} \\
	School of Computation, Information and Technology\\
	Technical University of Munich\\
	Munich, Germany \\
    \And
    \hspace{1mm}{Eleni D. Koronaki} \\
	Faculty of Science, Technology and Medicine\\
	University of Luxembourg\\
	Esch-sur-Alzette, Luxembourg \\
     \And
    \hspace{1mm}{Ronen Talmon} \\
	Viterbi Faculty of Electrical Engineering, Technion\\
	Israel Institute of Technology\\
	Haifa, Israel \\
     \And
    \hspace{1mm}{Ronald R. Coifman} \\
	School of Engineering $\&$ Applied Science\\
	Yale University\\
	New Haven CT, USA \\
     \And
    \hspace{1mm}{Erik Bollt} \\
	Electrical $\&$ Computer Engineering\\
	Clarkson University\\
	Potsdam NY, USA \\
    \And
    \hspace{1mm}{Ioannis G. Kevrekidis}\thanks{Correspondong author: yannisk@jhu.edu} \\
	Department of Chemical and Biomolecular Engineering\\
    Department of Applied Mathematics and Statistics\\
    Department of Urology\\
	Johns Hopkins University\\
	Baltimore MD, USA \\}
	
\renewcommand{\shorttitle}{On learning what to learn}

\begin{document}
\maketitle

\begin{abstract}
	Before we attempt to (approximately) learn a function between two (sets of) observables of a physical process, we must first decide what the {\em inputs} and what the {\em outputs} of the desired function are going to be. Here we demonstrate two distinct, data-driven ways of first deciding ``the right quantities'' to relate through such a function, and then proceeding to learn it.  This is accomplished by first processing multiple simultaneous heterogeneous data streams (ensembles of time series) from observations of a physical system: records of multiple {\em observation processes} of the system. We thus determine (a) what subsets of observables are {\em common} between the observation processes (and therefore observable from each other, relatable through a function); and (b) what information is {\em unrelated} to these common observables, and therefore particular to each observation process, and not contributing to the desired function. Any data-driven function approximation technique can subsequently be used to learn the input-output relation---from k-nearest neighbors and Geometric Harmonics to Gaussian Processes and Neural Networks. Two particular ``twists'' of the approach are discussed. The first has to do with the {\em identifiability} of particular quantities of interest from the measurements. We now construct mappings from {\em a single} set of observations from one process to {\em entire level sets} of measurements of the second process, consistent with this single set.  The second attempts to relate our framework to a form of causality: if one of the observation processes measures ``now'', while the second observation process measures ``in the future'', the function to be learned among what is common across observation processes constitutes a dynamical model for the system evolution.
\end{abstract}
\section{Introduction}
\label{intro}
In recent years the technology for observing/measuring phenomena and dynamic behavior in many disciplines, from physics and chemistry to biology and the medical sciences, has been growing at a spectacular pace – both the types of possible measurements as well as their spatiotemporal resolution and accuracy are constantly enriched.
It becomes thus increasingly possible to have several different measurements of the same phenomenon, observed simultaneously through different instruments (one could, for example, measure the extent of a reaction through measuring reactant/product concentrations or through measuring a physical property --say, a refractive index-- of the reacting mixture.)

\begin{figure}[h!]
    \centering
    \includegraphics[width=\ocf]{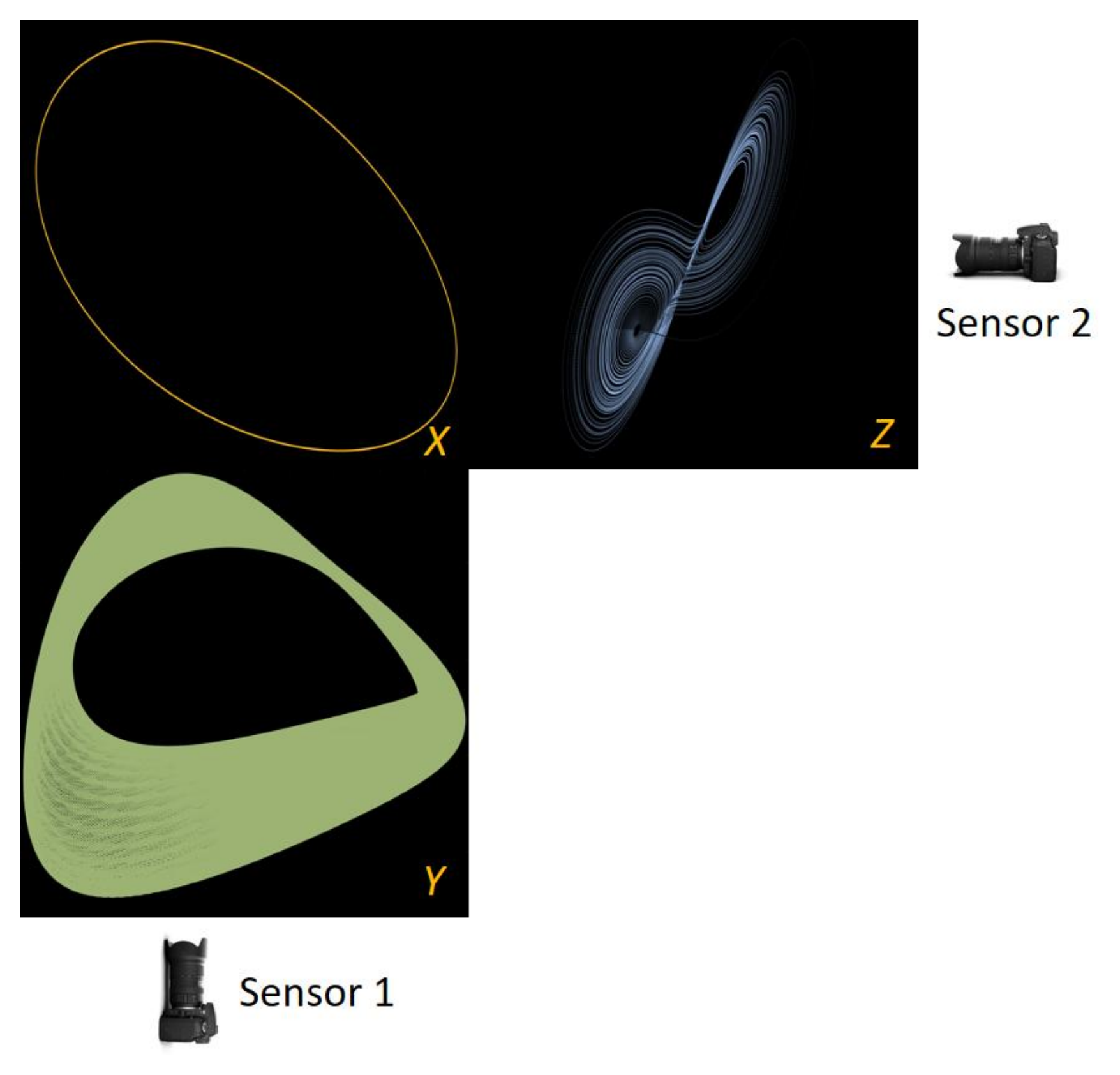}
    \caption{\label{fig:sensors} Illustrative Sensor setup: Sensor 1 only observes parts of systems X and Y. Sensor 2 only observes parts of systems X and Z.}
\end{figure}

The simultaneous progress in the mathematics of algorithms for data mining also open the way to registering such disparate measurements, and even fusing them. Discussions of ``gauge invariant data mining''~\cite{drira-2015,kemeth-2017a,kemeth-2018,haan-2020,kemeth-2022,yang-2023a}, 
that is, data mining that ultimately does not depend on the measuring instrument (as long as sufficiently rich information is collected) is a topic of active current research~\cite{singer-2008,singer-2009,talmon2013empirical,dsilva2013nonlinear}.
%
The ability to sufficiently accurately record the covariance of measurement noise around each measurement point is known to enable powerful tools for data registration/fusion~\cite{singer-2008,singer-2009,talmon2013empirical,dsilva2013nonlinear,dietrich-2020,moosmuller-2020,peterfreund-2020,gavish2022optimal,peterfreund-2023}. Different measurements of the same phenomenon (by which we imply measurements by different measuring instruments/observations through different observation functions) are often contaminated by instrument-specific distortion that hinders the registration/fusion task. 
This distortion could be {\em instrument specific noise}; alternatively (and the examples in this paper are based on this latter paradigm) each instrument may pick up, in addition to the process of interest, information from additional, unrelated processes, that take place ``in the vicinity'' of the measurement of interest. 
In the simplest case, Instrument 1 observes features of the ``process of interest'' X, as well as features of a single additional unrelated process (say Y); while Instrument 2 observes possibly the same or even different features of the same ``process of interest'' X, as well as features of an additional unrelated process, say Z, different from Y. This setup, involving measurements from two different Sensors, is introduced in Fig.\ref{fig:sensors}, and discussed in detail later in the manuscript.

The paradigm is directly motivated from the important relevant work of Lederman and Talmon\cite{lederman2018learning,katz2019alternating}, who used two cameras to observe three ``dancing'' robots (see Fig. \ref{fig:caricature} in the Appendix); one camera observed Yoda (Y) and the Bulldog (X), while another camera observed the Bunny (Z) and a {\em different view} of the Bulldog (a different observation of X). This paradigm has the additional convenience that the images of each robot in each camera do not overlap and therefore ``do not interact'': the ``measurement channels'' (the pixels of each camera) are what we will call ``{\em clean pixels}'' – they pertain to either the ``common process'' (the Bulldog) or to the particular camera's ``extraneous processes'' (Yoda or the Bunny).
The main result in Refs. \cite{lederman2018learning,katz2019alternating} was the development of an algorithm (the ``Alternating Diffusion'' algorithm) that jointly processes the data from both sensor streams, and discovers a data driven parametrization {\em of the common features across the sensors} (the measurements of the Bulldog). Here we will use their computational technology as the basis for {\em learning functions relating measurements of one camera to measurements of the other camera}. That is, we will construct --when possible-- observers of features measured by one sensor from features measured by the other sensor.


We will also briefly introduce and demonstrate another, more recently developed, algorithm for the extraction of so-called Jointly Smooth Functions (JSF, \cite{Dietrich2022}) as an alternative approach to the parametrization of the common features across two sensors (and therefore, as an alternative basis for learning cross-sensor observer functions). 
Other possible approaches to construct representations for common (and uncommon) coordinates between datasets are currently being explored. Coifman, Mashall, and Steinerberger~\cite{coifman-2023} propose a framework to identify such coordinates across graphs, while Shnitzer et al.~\cite{shnitzer-2019} propose anti-symmetric operator approximation to encode commonalities and differences. The latter has recently been extended by incorporating the Riemannian geometry of symmetric positive definite (SPD) matrices~\cite{katz2020spectral,shnitzer2024spatiotemporal}. 
In our work we go exploit the results that our two approaches (as well as these latter ones) can extract, to learn functional relations (observers) across different observations of the same dynamical system. 

We apply our two computational approaches to three distinct sets of nonlinear ordinary differential equations (our systems X, Y, and Z), observed from two sets of sensors: Sensor 1 observes time series of variables in systems X and Y, while Sensor 2 observes time series of variables in systems X and Z (hence, the common variables in our example pertain to the states of system X). Section \ref{sec:illustrative example} describes the systems we consider: (X) an autonomous limit cycle (periodic~\cite{Takoudis1981}), (Y) a periodically forced oscillator system (resulting in quasiperiodic dynamics~\cite{McKarnin1988}), and (Z) the Lorenz system~\cite{lorenz-1963} constrained to its (chaotic) attractor.

In section \ref{sec:learning functions} (also see Appendix~\ref{app:learning functions}), we show how the data-driven parametrization of common features across sensors ``discover'' which sets of individual Sensor 2 channels can be written as functions of some subset of Sensor 1 channels, and {\em vice versa}. We demonstrate this learning process using several commonly available alternative methods: k-nearest neighbors (KNN~\cite{fix-1951}), geometric harmonics (GH~\cite{Coifman2006a,dietrich-2021a}), and feed-forward neural networks (FFNN~\cite{KingmaB14}).

Having demonstrated the base case, later sections discuss potential problems and extensions. Section \ref{sec:mixed channels} discusses the case where individual sensor channels \emph{do not} belong to observations of a single system (what we called ``clean pixels''), but rather constitute a combination of observations of multiple systems (what we call ``dirty pixels''). Specifically, we apply random linear transformations to each set of sensor data, so that each individual sensor channel variable is a linear combination of all measured variables from that sensor's two relevant systems. Even in this more challenging setting, our computational approach can extract that system X is commonly observed by both sensors. In this case, one sensor's observations cannot predict any particular channel of the other sensor; the second sensor channels are ``unidentifiable'' from measurements of the first sensor. Instead, we can describe {\em a level set} of the second sensor's full measurement space that is {\em consistent
with} the particular observations of the first sensor. We discuss how to parameterize such level sets using a manifold learning variant called {\em Output-Informed} Diffusion Maps~\cite{Lafon2004,holiday-2019}.

In Section \ref{sec:temporal}, we consider the case when the channel measurements from Sensor 2 include ``future'' measurements of variables measured ``now'' by Sensor 1. This allows us to learn approximate evolution equations for the system that is common between the two sensors, establishing a certain type of causality between the two sets of measurements. 

We conclude with further thoughts on the parameterization of the``uncommon variable'' level sets, including the observation of common/uncommon features across scales, the possible use of new, conformal  neural network architectures for this purpose, as well as good sampling techniques on these ``uncommon'' level sets.

\section{ILLUSTRATIVE EXAMPLES}
\label{sec:illustrative example}

\subsection{\bf Models of a periodic (X), a quasiperiodic (Y) and a chaotic (Z) response.}

To illustrate how manifold learning leads to finding common features across different sensor measurements and learning relations between them,  we generated data from three independent nonlinear dynamical systems. For our common process $X$, we will use data from a surface reaction model studied by Takoudis et. al.\cite{Takoudis1981}, which modifies the Langmuir-Hinshelwood mechanism by requiring two empty surface sites in the surface reaction step:
\begin{equation}
    \begin{gathered}
        A+B\rightleftharpoons AS, \\
        B+S\rightleftharpoons BS, \\
        AS + BS + 2S \rightleftharpoons 4S + \text{products}
    \end{gathered}
\end{equation}
After non-dimensionalizing the rate equations, we obtain a system of two nonlinear differential equations in $\theta_A$ and $\theta_B$, the fractional surface coverages of the two reactants,
and four parameters:
\begin{equation}
    \begin{gathered}
        \frac{d\theta_A}{dt}=\alpha_1(1-\theta_A-\theta_B)-\gamma_1\theta_A-\theta_A\theta_B(1-\theta_A-\theta_B)^2, \\
        \frac{d\theta_B}{dt}=\alpha_2(1-\theta_A-\theta_B)-\gamma_2\theta_B-\theta_A\theta_B(1-\theta_A-\theta_B)^2.
    \end{gathered}
\end{equation}
This system exhibits sustained oscillations for certain parameter values; we
will sample data from the limit cycle arising for $\gamma_{1} = 0.001$, $\gamma_{2} = 0.002$, $\alpha_{1} = 0.016$, $\alpha_{2} = 0.0278$.

For our first sensor-specific process $Y$¸ we will use data from a periodically forced version of the above oscillatory system:
%
a forcing term with the nondimensional form
\begin{equation}
    \alpha_2 = A_0 + A\cos(\omega t).    
\end{equation}
is added, periodically perturbing the gas-phase pressure of B.
For $\gamma_{1} = 0.001$, $\gamma_{2} = 0.002$, $\alpha_{1} = 0.019$, $A_{0} = 0.028$, $A = 0.002097$, $\omega = 0.01722$,
the long-term dynamics are  quasiperiodic~\cite{McKarnin1988}.

For our second sensor-specific process $Z$, we will use data generated on the attractor of the Lorenz system~\cite{lorenz-1963},
\begin{equation}
        \frac{dx}{dt}=\sigma(y-x), 
        \frac{dy}{dt}=x(\rho-z)-y, 
        \frac{dz}{dt}=xy-\beta z.
\end{equation}
We use the parameter value set $\sigma=10$, $\beta=\frac{8}{3}$, $\rho=28$, which is known to result in chaotic dynamics.

We define our sensor setup so that the first sensor can only detect time series data of the variable $\theta^X_A$ from system $X$, and also of the variable with the same name, $\theta^Y_A$, from system $Y$.
The second sensor can only detect time series of $\theta^{(X)}_B$ from system $X$ and $y$ from system $Z$. 
We include a time-delayed measurement for each channel, so that we can fully capture the dynamics of the common (periodic) system (in the spirit of Whitney~\cite{whitney-1936,sauer-1991} and Takens~\cite{ruelle-1971,takens-1981,stark-1997,moosmuller-2020}), see Fig.~\ref{fig:sensors}:

\begin{equation}
    \begin{gathered}
        S^{(1)}(t) = \lbrack\theta_{A}^{(X)}\left( t \right),\theta_{A}^{(Y)}\left( t \right),\theta_{A}^{(X)}\left( t - \Delta t \right),\theta_{A}^{(Y)}\left( t - \Delta t \right)\rbrack, \\
        S^{(2)}(t) = \lbrack\theta_{B}^{(X)}\left( t \right),y\left( t \right),\theta_{B}^{(X)}\left( t - \Delta t \right),\ y\left( t - \Delta t \right)\rbrack. 
    \end{gathered}
\end{equation}
We take simultaneous measurements from each sensor at a sampling rate sufficiently faster than the frequency of our common system; due to their different frequencies/ different natures of the responses, each system's measurements cannot be long-term correlated with measurements of the other two systems.

The computational tools that will be used to process the data from these numerical experiments are discussed in Appendix~\ref{sec:computational methods}. They include Diffusion Maps (and Output-informed Diffusion Maps), Alternating Diffusion, Jointly Smooth Function extraction, and Local Linear Regression (LLR).
The techniques for learning functions as a post-processing of the data analysis include k-nearest neighbors (KNN), Geometric Harmonics (GH), and ``vanilla'' (Multilayer Perceptron) Feed-Forward Neural Networks (FFNN). The corresponding algorithms are included in Appendix~\ref{app:learning functions}.

\subsection{Alternating-Diffusion Embedding}

\begin{figure}
    \includegraphics[width=0.5\textwidth]{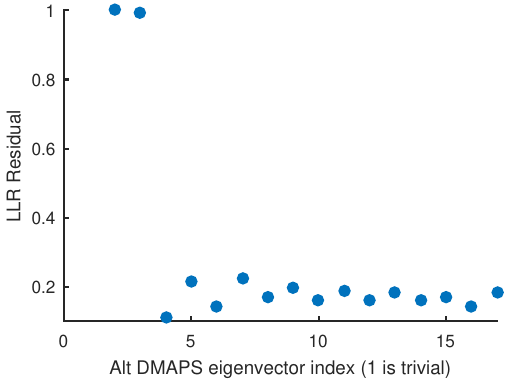}
    \centering
    \caption{\label{fig:basic_LLR} Results of running LLR on the set of successive alternating-diffusion eigenvectors $\phi_i$ (sorted by eigenvalue). $\phi_1$ is trivially constant, and $\phi_2$ has a normalized LLR residual of 1 by definition. $\phi_2$ is the only other top eigenvector with a high residual, indicating that it represents a unique direction and that the most parsimonious embedding of the common system is two-dimensional.}
\end{figure}

\begin{figure*}
    \includegraphics[width=0.7\textwidth]{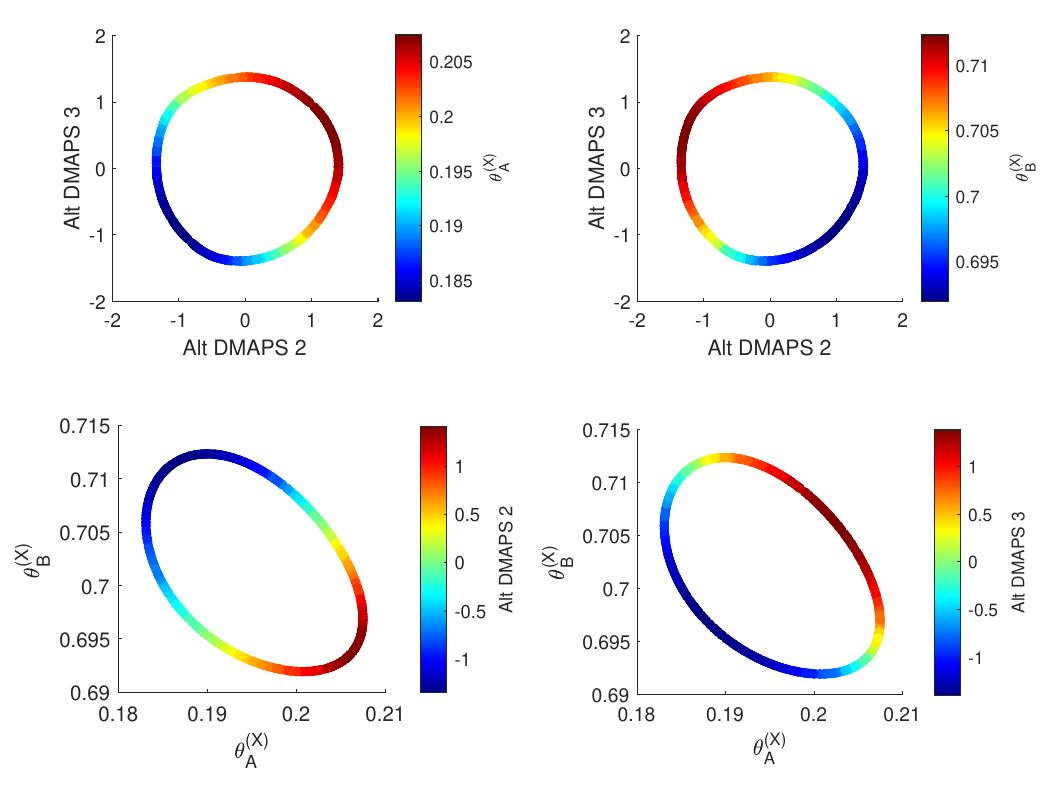}
    \centering
    \caption{\label{fig:basic_embed} These plots confirm that the alternating-diffusion embedding is one-to-one/bi-Lipschitz with the coordinates of the common system X. (top) Plots of the alternating-diffusion embedding colored by $\theta_A^{(X)}$ (left) and $\theta_B^{(X)}$ (right). (bottom) Plots of $\theta_B^{(X)}$ vs. $\theta_A^{(X)}$, colored by alternating-diffusion eigenvectors 2 (left) and 3 (right).}
\end{figure*}

We constructed our alternating-diffusion operator~\cite{lederman2018learning} as the product of two diffusion operators, each based on the Euclidean distances of the observations of Sensor 1 and, separately, of Sensor 2. We used LLR to analyze the true dimensionality of the recovered common coordinates and found that the first two non-trivial Alternating Diffusion eigenvectors represented unique coordinates (see Fig. \ref{fig:basic_LLR}). In Fig. \ref{fig:basic_embed}, we can visually confirm that the recovered common coordinates are one-to-one/bi-Lipschitz with the coordinates of the common system X.

\begin{figure*}
    \includegraphics[width=\tcf]{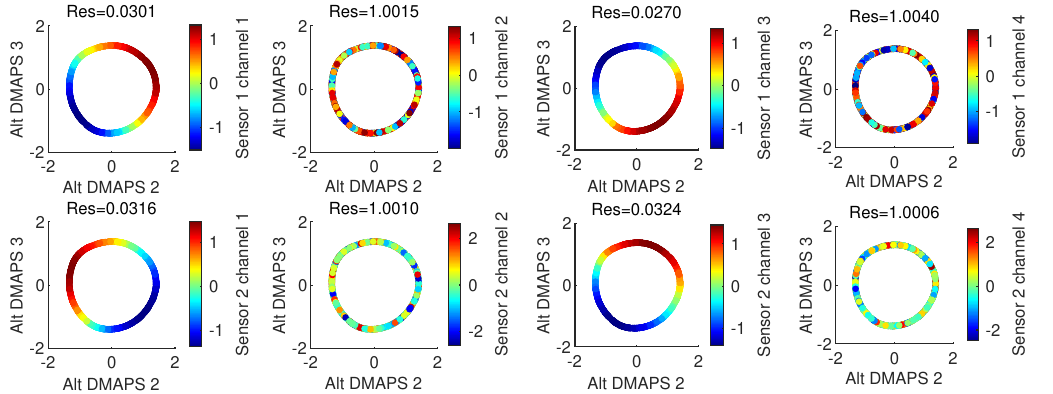}
    \caption{\label{fig:basic_coord} Plots of the alternating-diffusion embedding colored by each of the individual sensor channels, with the LLR residual above each plot. Channels 1--4 of Sensor 1 (top row) are the measurements $\lbrack\theta_{A}^{(X)}\left( t \right),\theta_{A}^{(Y)}\left( t \right),\theta_{A}^{(X)}\left( t - \Delta t \right),\theta_{A}^{(Y)}\left( t - \Delta t \right)\rbrack$, while channels 1--4 of Sensor 2 (bottom row) are the measurements $\lbrack\theta_{B}^{(X)}\left( t \right),y\left( t \right),\theta_{B}^{(X)}\left( t - \Delta t \right),\ y\left( t - \Delta t \right)\rbrack$. Coordinates that belong to the common system (Sensor 1 channels 1 and 3, Sensor 2 channels 1 and 3) have a low residual and appear visually smooth. Other coordinates have a high residual and appear noisy.}
\end{figure*}

In general, Alternating Diffusion does not require that each sensor channel (each camera pixel) involves observations of just one system (what we called ``clean'' channels or ``clean'' pixels above).
%
Sensor channels that {\em combine} simultaneous measurements from the common system and one or more sensor-specific systems (what we call ``dirty'' channels or ``dirty'' pixels) cannot therefore be written as a function of our Alternating Diffusion common coordinates (they are not \textit{identifiable} from these common coordinates). 
However, in this current section, we will consider the case where (at least some) of our original sensor observations are ``clean'', i.e., they only relate to our common system. We identify these ``clean'' channels using LLR (see Fig. \ref{fig:basic_coord}). 
Later, in section~\ref{sec:mixed channels}, we will also demonstrate how to extract {\em common} as well as {\em uncommon} coordinates if there are no ``clean'' observations available.
Note also in Fig~\ref{fig:basic_coord} (col. two and four) that sensor-specific observations (pixels) {\em are not smooth functions of the common coordinates} (their Dirichlet energy appears visually extremely high).
These measurements are clearly  \textit{not identifiable} from the common coordinates.

\subsection{Jointly Smooth Functions}

\begin{figure*} 
\centering
\begin{tabular}{cc}
\includegraphics[width=0.6\textwidth]{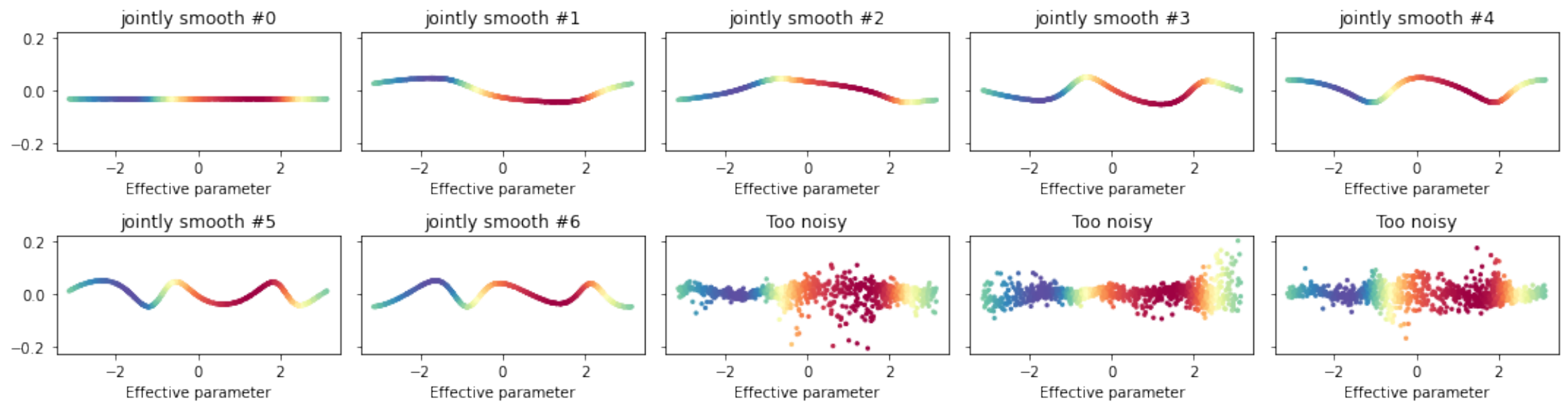} & 
\includegraphics[width=0.4\textwidth]{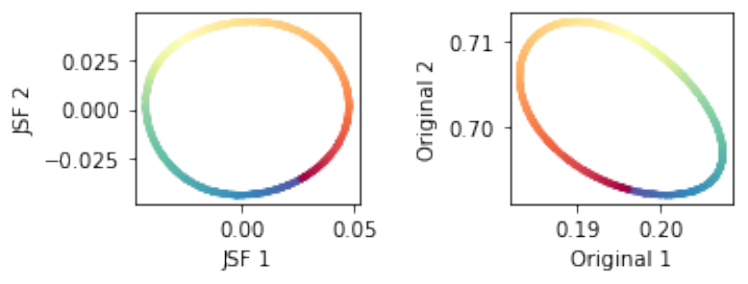} \\
    (a) & (b) 
\end{tabular}

\caption{(a) The first 10 extracted jointly smooth functions. (b)(Left)The embedding result for the two most parsimonious JSFs. \textit{(Right)} The original system X data colored by one JSF.}
\label{fig:function map results}
\end{figure*}

We apply Jointly Smooth Functions (JSF)~\cite{Dietrich2022} to the same sensor data described above. In Fig. \ref{fig:function map results}a, we visualize the first ten JSF. As we can observe, only the first seven JSF are smooth (have low Dirichlet energy). Similarly to Alternating Diffusion, we can use LLR to select the two functions which give the most parsimonious embedding: they are the second and third JSF (\#1 and \#2), whose relative shift is  reminiscent of the shift between a sine and a cosine function.
The common system coordinates are ``nice'' (low Dirichlet energy) functions of the chosen JSF (see Fig.~\ref{fig:function map results}b).


\section{Learning functions across sensors}\label{sec:learning functions}
\begin{figure*} 
\centering
\begin{tabular}{ccc}
    \includegraphics[width=0.3\textwidth]{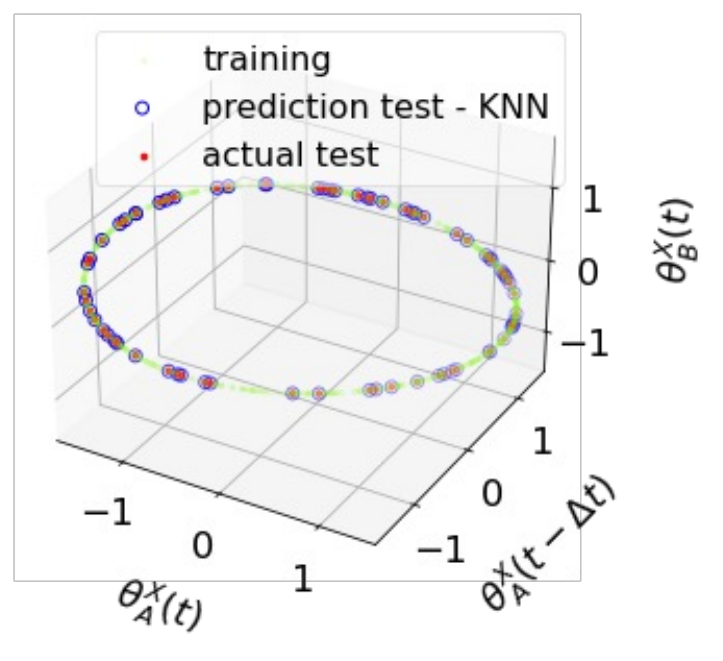} &  \includegraphics[width=0.3\textwidth]{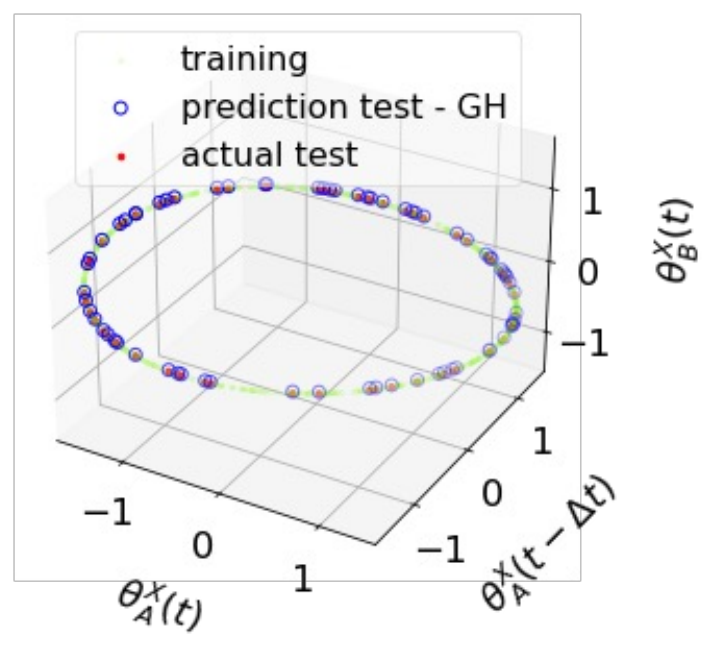} & 
\includegraphics[width=0.3\textwidth]{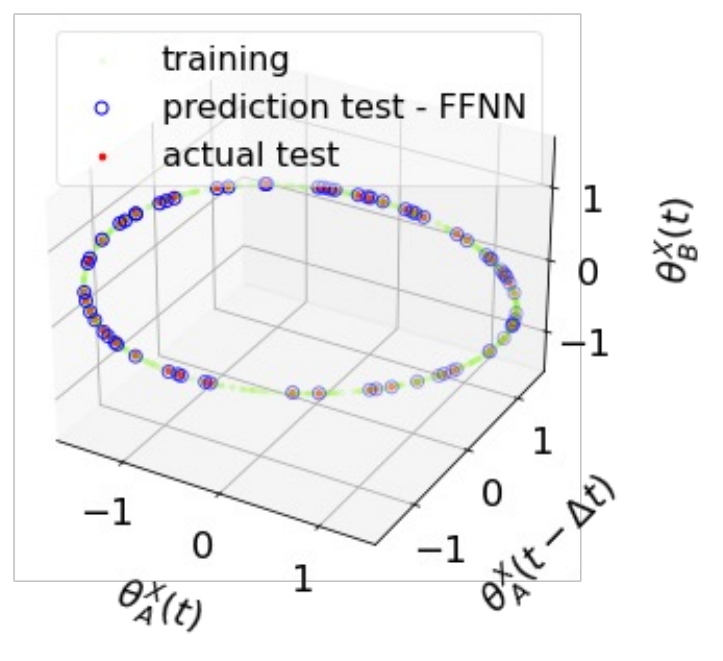} \\
    (a) & (b) & (c) 
\end{tabular}

\caption{(a) The predicted values of $\theta_B(t)$ in blue compared with the true values in red for testing data, as well as the labeled (training) points in green, using KNN. (b) Using GH to learn the function. (c) Using FFNN to learn the function.}
\label{fig:function map results1}
\end{figure*}

Once we have found which measurements of one sensor stream 
``belong together'' with which measurements of the second sensor stream, {\em through their joint parametrization by common features},
we can approximate, in a data-driven manner, the relation between them.
In this section we describe several approaches for achieving this function approximation: nearest neighbor search, geometric harmonics, and artificial neural networks. We demonstrate these methods on a dataset of samples including $ (\theta^X_A(t)$, $\theta^X_A(t-\Delta t)$, $\theta^X_B(t))$.
The first two coordinates are seen by Sensor 1, and are one-to-one with the identified common coordinates. The last one is seen by Sensor 2, and is a ``clean'' channel measurement: it should be possible to learn $\theta^X_B(t)$ as a function of the first two, i.e., $(\theta^X_A(t)$, $\theta^X_A(t-\Delta t)$.
The dataset is split in training and testing subsets. For the training set, the values of $\theta^X_A(t)$, $\theta^X_A(t-\Delta t)$, $\theta^X_B(t)$ are known, while for the test set, we have only the values of $\theta^X_A(t)$, $\theta^X_A(t-\Delta t)$ and we will ``predict'' or ``fill in''  $\theta^X_B(t)$ values.
For this section, we have used the first 50 sample data points for our function learning algorithms.
The accuracy of each function learning algorithm is quantified based on the $L_{\infty}$ norm for $n_{samples}=200$ values of $\theta_B(t)$,
\begin{equation}
\label{eq:error}
\varepsilon = \frac{||\theta_B(t)_{true}-
\theta_B(t)_{predicted}||_{\infty}}{n_{samples}}.
\end{equation}
Fig.~\ref{fig:function map results1}(a-c) shows the results when using KNN (a), GH (b), and FFNN (c) to map from two measurements of the common system X (measured by Sensor 1) to a ``clean'' measurement of the common system X measured by Sensor 2. All methods of approximation produce accurate extrapolation results on the limit cycle.
\begin{figure*}[htp!]
    \includegraphics[width=\tcf]{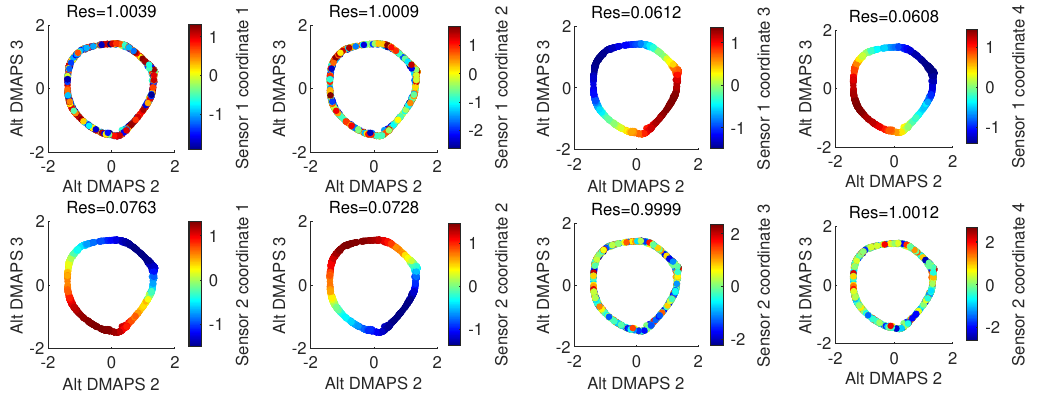}
    \caption{\label{fig:caus1_coord_check} For the first setup, plots of the alternating-diffusion embedding colored by each of the individual sensor channels, with the LLR residual above each plot. Channels 1--4 of Sensor 1 (top row) are the measurements $\lbrack\theta_{A}^{(X)}\left( t \right),\theta_{A}^{(Y)}\left( t \right),\theta_{A}^{(X)}\left( t - \Delta t \right),\theta_{A}^{(Y)}\left( t - \Delta t \right)\rbrack$, while channels 1--4 of Sensor 2 (bottom row) are the measurements $\lbrack\theta_{B}^{(X)}\left( t \right),y\left( t \right),\theta_{B}^{(X)}\left( t - \Delta t \right),\ y\left( t - \Delta t \right)\rbrack$.  Coordinates that belong to the common system (Sensor 1 channels 3 and 4, Sensor 2 channels 1 and 2) have a low residual and appear visually smooth. Other coordinates have a high residual and appear noisy.}
\end{figure*}

\section{Learning Causality}
\label{sec:temporal}


Given the computational tools demonstrated in this work so far, we are now faced with an interesting possibility: if Sensor 1 gives us measurements ``now'' and Sensor 2 gives us measurements of the same quantities ``in the future'', our common coordinates will allow us to learn quantities in the future \textit{as a function of the same quantities now} - that is, help us learn a dynamical model of the common process. This brings us close to the idea (and the entire field) of data-driven causality. 

A most basic premise to questions of causation is the principle that cause comes before the effect, but furthermore, a causal influence is one where the outcome is related to its cause.  As simple as this concept may seem, it becomes nontrivial to develop a definition that is both robust but also testable in terms of data and observations. 
Two major schools of thought have arisen in modern parlance: the perspective of information flow, and the perspective of interventions.  The information flow perspective includes the Nobel prize winning work on Granger-causality \cite{granger1969investigating}, and the recently highly popular transfer entropy \cite{schreiber2000measuring} (TE), causation entropy \cite{sun2014causation, sun2015causal, sun2014identifying,surasinghe2020geometry} (CSE), Cross Correlation Method (CCM) \cite{sugihara2012detecting}, Kleeman-Liang formalism \cite{liang2018causation} and others, these being probabilistic in nature. In some sense these all address the question of whether an outcome $x$ is better forecast by considering an input variable $y$ at a previous time, or not.  If yes, then $y$ is considered causal.  However, the \textit{intervention} concept, most notably developed in the ``Do-calculus'' of Pearl \cite{pearl2009causality}, is premised on a formalism of interventions and counterfactuals that are typically decided with data in terms of a specialized Bayesian analysis.

With the concept of common variables described in this paper, we are presented with the possibility of a different path to define causal relationships  by asking the simple question as to whether observations of certain variables in the past are ``common'' with (contain sufficient common information to predict) observation of these variables in the future.
By the data-driven methods developed here, we need only to prepare the data in the following manner:  assume a stochastic process produces a sequence of vector valued data, $\{ {\bf x}(t_i) \}_{i=s_1}^{s_2}$. Also, let ${s_1, \dots, s_2}$ be a discrete index set, and ${\bf x}(t):{\mathbb R}\rightarrow {\mathbb R}^d$.  In our wording, Sensor 1 is shown multiple instances of past vector observations,  ${\mathcal X}=\{ {\bf x}(t_i)\}$ and Sensor 2 is shown multiple instances of the corresponding future observations ${\mathcal X}’=\{ {\bf x}(t_{i+1})\}$. Then the ``common'' coordinates connecting past and future may be understood as having a casual relationship. 
In these terms, clean observations of the common system by Sensor 1 (now) are causally related to clean observations of the same common system variables by Sensor 2 (the future):  there exists a data-driven scheme that develops a nontrivial functional relationship from past observations to future outcomes. 
We are avoiding the phrase ``correlate'' because that has statistical connotations, usually assuming a linear relationship. Our common coordinate-based mapping from the present to the future is a {\em deterministic}, nonlinear one.
%
%
Furthermore, while this machine learning/manifold-learning based approach is distinct from the DO-calculus  there may exist a path to connect them: a bridge could be conceptually constructed if the data set itself includes some parametric interventions.  
Otherwise, it has aspects common to the Wiener-Granger causality concept of forecastability.
%

 
In our first setup, Sensor 1 sees $\theta^X_A(t)$ and $\theta^X_B(t)$ from system X, and $\theta^Y_A(t)$ and $\theta^Y_B(t)$ from system Y. Sensor 2 sees $\theta^X_A(t + \tau)$ and $\theta^X_B(t + \tau)$ from system X, and $x(t + \tau)$ and $y(t + \tau)$ from system Z, where here $\tau = 200$ time units, about 25\% of the period of system X. By using time-shifted measurements, Sensor 2 effectively sees ``into the future'' of system X, which will allow us to approximate the evolution equations for the system X variables.

\begin{figure}[ht]
    \includegraphics{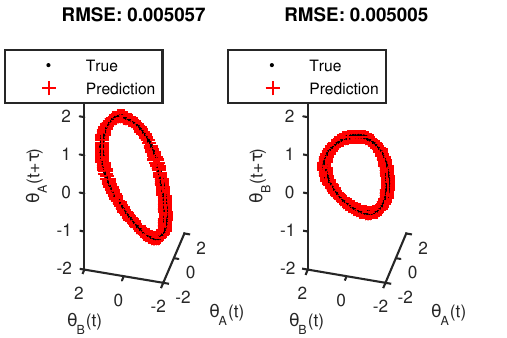}
    \centering
    \caption{\label{fig:caus1_learning} Learning causality with KNN. Here, we learn a map from $(\theta^X_A(t),\theta^X_B(t))$ (the present) to $(\theta^X_A(t+\tau),\theta^X_B(t+\tau))$ (the future). }
\end{figure}

We use the LLR algorithm to determine that the alternating diffusion embedding for this sensor setup is two-dimensional. Visually, and the with LLR algorithm, we can determine which observables from each sensor are related to system X. In Fig. \ref{fig:caus1_coord_check},
the title shows the normalized residual value from the LLR algorithm. The variables which have residuals close to 0 are functions of the alternating diffusion embedding, and thus can be assumed to only be related to system X. We can then learn functions from $[\theta_A(t),\theta_B(t)]$ to $\theta_A(t + \tau)$ and $\theta_B(t + \tau)$, effectively approximating the evolution equations. For example, the results from learning using a 5 nearest neighbors regression are shown in Fig. \ref{fig:caus1_learning}. 

We can also apply Jointly Smooth Functions to the same sensor data described above. The results are presented in \textcolor{black}{Appendix} \ref{app:JSFres}.

\begin{figure*}
    \includegraphics[width=\tcf]{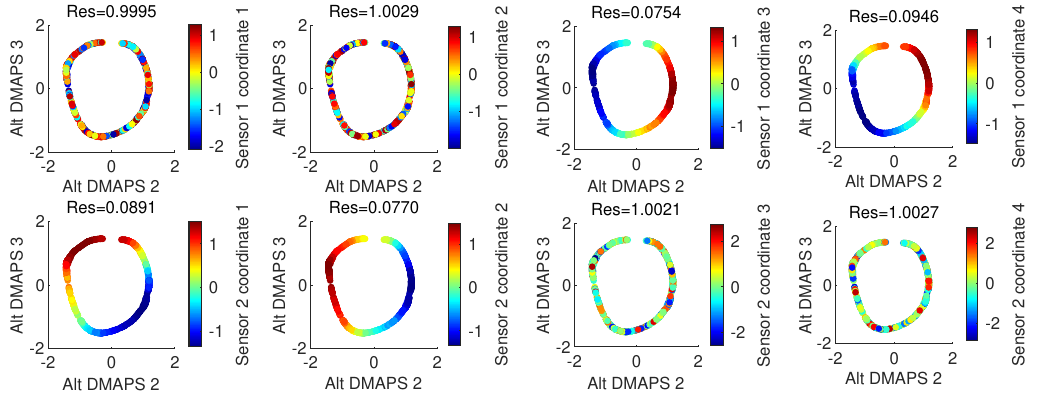}
    \caption{\label{fig:caus2_coord_check} For the second setup, plots of the alternating-diffusion embedding colored by each of the individual sensor channels, with the LLR residual above each plot.  Channels 1--4 of Sensor 1 (top row) are the measurements $\lbrack\theta_{A}^{(X)}\left( t \right),\theta_{A}^{(Y)}\left( t \right),\theta_{A}^{(X)}\left( t - \Delta t \right),\theta_{A}^{(Y)}\left( t - \Delta t \right)\rbrack$, while channels 1--4 of Sensor 2 (bottom row) are the measurements $\lbrack\theta_{B}^{(X)}\left( t \right),y\left( t \right),\theta_{B}^{(X)}\left( t - \Delta t \right),\ y\left( t - \Delta t \right)\rbrack$.  Coordinates that belong to the common system (Sensor 1 channels 3 and 4, Sensor 2 channels 1 and 2) have a low residual and appear visually smooth. Other coordinates have a high residual and appear noisy.}
\end{figure*}

\begin{figure} [h!]
    \includegraphics{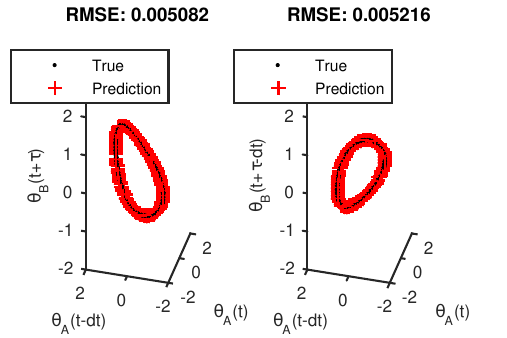}
    \centering
    \caption{\label{fig:caus2_learning} Functions from $[\theta^X_A(t),\theta^X_A(t-\Delta t)]$ to $\theta^X_B(t + \tau)$ and $\theta^X_B(t+\tau-\Delta t)$, constructed with a five-nearest neighbors regression.}
\end{figure}

For our second setup, Sensor 1 sees $\theta^X_A(t)$ and $\theta^X_A(t-\Delta t)$ from system X, and $\theta^Y_A(t)$ and $\theta^Y_A(t-\Delta t)$ from system Y. Sensor 2 sees $\theta^X_B(t+\tau)$ and $\theta^X_B(t+\tau-\Delta t)$ from system X and $y(t+\tau)$ and $y(t+\tau-\Delta t)$ from system Z. Here, $\Delta t = 100$ time units and $\tau=250$ time units. Visually, and with the LLR algorithm, we show which observables from each sensor are related to each other (Fig.~\ref{fig:caus2_coord_check}). We can learn functions from $[\theta^X_A(t),\theta^X_A(t-\Delta t)]$ to $\theta^X_B(t + \tau)$ and $\theta^X_B(t+\tau-\Delta t)$ with a five-nearest neighbors regression (Fig.  \ref{fig:caus2_learning}).

We can apply jointly smooth functions to the same sensor data described above and the results are shown in Appendix \ref{app:JSFres}.

\section{Mixed sensor channels}
\label{sec:mixed channels}

What if our sensor measurement channels are ``dirty'', meaning they involve combinations of measurements from the common and the sensor-specific observations?
In this section, we apply the Alternating Diffusion framework to sets of sensor data that are not directly separable into common and uncommon parts. All observations of each sensor are influenced by both the common and the sensor-specific system.
Even in this setting, Alternating Diffusion correctly uncovers a parametrization of the common system. 

\subsection{Application to the oscillatory reaction example}

Here the measurements of Sensor 1 are linear combinations of {\em all} the ``clean'' Sensor 1 channels -- and the same thing holds for the measurements of Sensor 2.
%


Beyond the mixing of the Sensor 1 measurements, we use --for Sensor 2  measurements time-shifted by a fixed amount (approximately 25\% of the period of system 2) {\em and} take linear combinations of them.

More explicitly, the sensor measurements are given by
\begin{equation}
\begin{gathered}
S^{(1)}=
\begin{tabular}{|l|l|l|l|}
\hline $\theta_A^{(X)}(t)$ & $\theta_B^{(X)}(t)$ & $\theta_A^{(Y)}(t)$ & $\theta_B^{(Y)}(t)$ \\
\hline 
\end{tabular}
\times
\begin{tabular}{|l|l|l|l|}
\hline   0.3637 &   -0.0173 &   -0.3701 &	 0.1013 \\
\hline  -0.5068 &   -0.4513	&    0.1470	&   -0.2041 \\
\hline   0.0888 &   -0.1818	&    0.3284	&    0.3344 \\
\hline   0.0407 &   -0.3496	&   -0.1545	&    0.3602 \\
\hline
\end{tabular}
 \\ \\
S^{(2)}=
\begin{tabular}{|l|l|l|l|}
\hline $\theta_A^{(X)}(t+dt)$ & $\theta_B^{(X)}(t+dt)$ & $x^{(Z)}(t+dt)$ & $y^{(Z)}(t+dt)$ \\
\hline 
\end{tabular}
\times
\begin{tabular}{|l|l|l|l|}
\hline  -0.1394 &   -0.0597 &    0.0828 &	 0.3847 \\
\hline  -0.3803 &   -0.3576	&   -0.3440	&   -0.0628 \\
\hline  -0.1010 &   -0.5259	&   -0.2147	&   -0.3981 \\
\hline  -0.3793 &   -0.0568	&    0.3585	&   -0.1544 \\
\hline
\end{tabular}.
\end{gathered}
\end{equation}

\begin{figure}
    \includegraphics[width=0.4\textwidth]{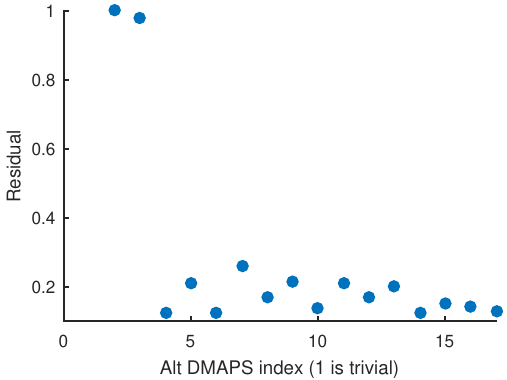}
    \centering
    \caption{\label{fig:output_LLR} Results of running LLR on the set of successive alternating-diffusion eigenvectors $\phi_i$ (sorted by eigenvalue). $\phi_1$ is trivially constant, and $\phi_2$ has a normalized LLR residual of 1 by definition. $\phi_2$ is the only other top eigenvector with a high residual, indicating that it represents a unique direction and that the most parsimonious embedding is two-dimensional.}
\end{figure}

\begin{figure}[ht]
    \includegraphics[width=\textwidth]{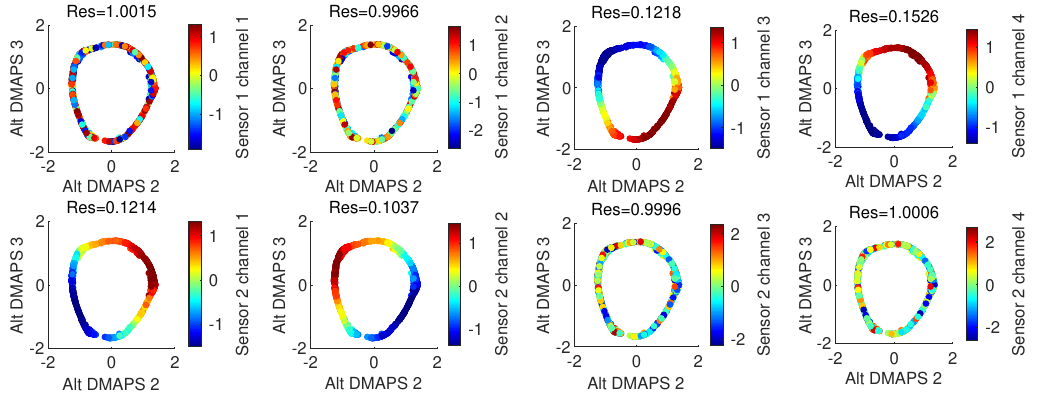}
    \caption{\label{fig:output_coord_check} Plots of the alternating-diffusion embedding colored by each of the {\em untransformed} individual sensor channels, with the LLR residual above each plot.  Channels 1--4 of Sensor 1 (top row) are the measurements $\lbrack\theta_{A}^{(X)}\left( t \right),\theta_{A}^{(Y)}\left( t \right),\theta_{A}^{(X)}\left( t - \Delta t \right),\theta_{A}^{(Y)}\left( t - \Delta t \right)\rbrack$, while channels 1--4 of Sensor 2 (bottom row) are the measurements $\lbrack\theta_{B}^{(X)}\left( t \right),y\left( t \right),\theta_{B}^{(X)}\left( t - \Delta t \right),\ y\left( t - \Delta t \right)\rbrack$.  Coordinates that belong to the common system (Sensor 1 channels 3 and 4, Sensor 2 channels 1 and 2) have a low residual and appear visually smooth. Other coordinates have a high residual and appear noisy.}
\end{figure}

Here, $\Delta t$ = 200 time units, about 25\% of the period of the common system.
The resulting alternating diffusion embedding is two-dimensional (Fig. \ref{fig:output_LLR}). Coloring the embedding by the {\em untransformed} X coordinates (Fig. \ref{fig:output_coord_check}) shows that we have indeed captured system X.

We can apply jointly smooth functions to the same sensor data described above. The results are presented in Appendix \ref{app:JSFres}.

\section{Output-informed Diffusion Maps}
\label{sec:outputDMAPs}
Even though the parametrization of the common system is discovered by either Alternating Diffusion Maps or Jointly Smooth Functions, we cannot, in this case, learn a function from the common Alternating Diffusion Maps (AltDmaps) embedding to any of the individual original sensor channels. We can only say that points with the same AltDmaps embedding value {\em will lie on a particular level set} in the original sensor observation space. So, if we know enough information from Sensor 1 to find where we are in the AltDmaps common embedding, we cannot tell what Sensor 2 will simultaneously measure - but we can tell {\em what level set the measurements from Sensor 2 will lie on}. 
Sensor 2 measurements are thus {\em structurally unidentifiable} from Sensor 1 measurements in this case. 
For example, if X and Y are limit cycles with different (irrationally related) periods, and if Sensor 1 measures at a particular phase of X, there will be many possible corresponding phases of Y -- a one-parameter family of them -- and they could be parameterized \textit{by an embedding of the \underline{uncommon system}}. To find this embedding, we can use a modification  of Diffusion Maps, the so-called {\em Output-Informed Diffusion Maps}~\cite{Lafon2004,holiday-2019}, presented briefly here for clarity.

The goal of output-informed diffusion maps is to parameterize manifolds when variation along some directions on the manifold produces no response in some output measurement. In a typical scenario, the input manifold will be a sampling of the space of parameters for some dynamical system, and the output measurement will be the time series response of the system variables. If some parameter combinations are redundant (e.g., if only the ratio of two parameters influences the system response), the output manifold will have a lower-dimensionality than the input manifold. We would like to separate our parameterization of the input manifold so that the leading coordinates impact the system response, and they are followed by coordinates that do not. To accomplish this, we introduce a new kernel (first proposed in a different context in the Thesis of S. Lafon \cite{Lafon2004}, and also used in a similar identifiability context in {\cite{holiday-2019}}): let $f(y_i)$ be the output response for input measurement $y_i$:
\begin{equation}
\label{output}
w(y_i,y_j)=\exp\left(- \frac{||f(y_i)-f(y_j)||^2}{\epsilon^2} - \frac{||y_i-y_j||^2}{\epsilon} \right).
\end{equation}
Since $\epsilon$ is typically less than one (or can be made so by scaling the original data), this kernel overemphasizes directions on the input manifold that actually result in changes in the output response.

In our case, we use the sensor data as the input manifold with the AltDmaps embedding as the ``output,'' which factors the standard Dmaps embedding of Sensor 1 into {\em common} and {\em uncommon} eigenvectors. This also then gives us an embedding of the uncommon system (and an understanding of its dimensionality), as well as coordinates which parameterize the common level sets.

\subsection{Application to the oscillatory reaction example}
\begin{figure*} 
\centering
\begin{tabular}{cc}
    \includegraphics[width=0.35\textwidth]{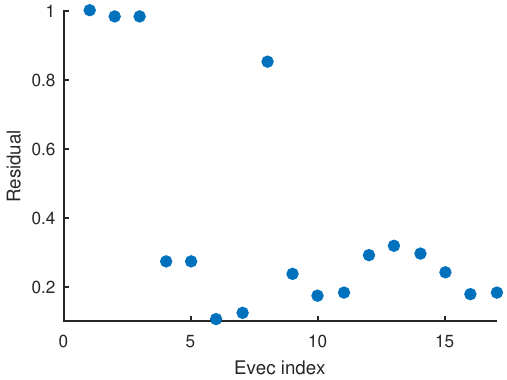} & 
\includegraphics[width=0.6\textwidth]{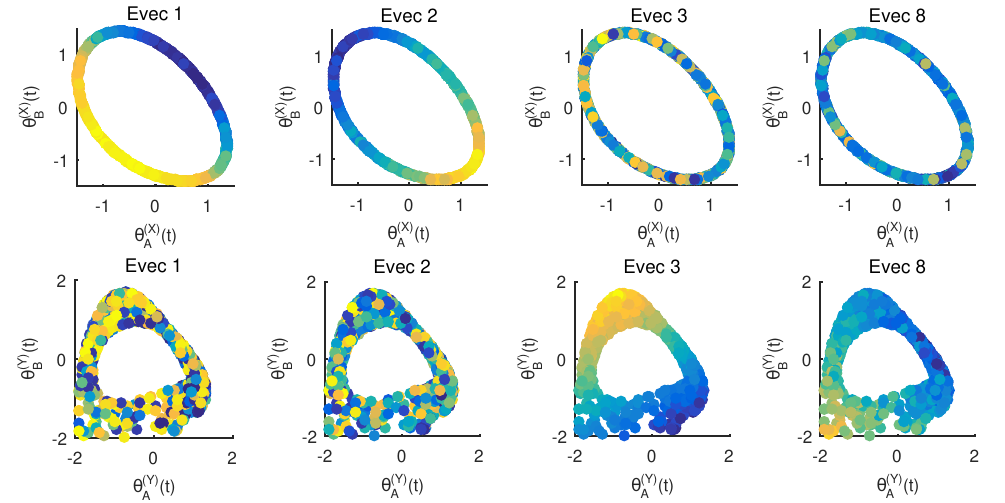} \\
    (a) & (b) 
\end{tabular}

\caption{(a) Results of running LLR on the set of successive eigenvectors $\phi_i$ (sorted by eigenvalue) from output diffusion maps on Sensor 1 data with the alternating-diffusion eigenvectors as the output. $\phi_1$ is trivially constant, and $\phi_2$ has a normalized LLR residual of 1 by definition. Eigenvectors 1, 2, 3, and 8 represent unique directions. (b)(Top row) Plots of the system X variables, colored by the output diffusion map eigenvectors 1, 2, 3, and 8. (Bottom row) Plots of the system Y variables, colored by the output diffusion map eigenvectors 1, 2, 3, and 8.}
\label{fig:output1}
\end{figure*}

We can now use the alternating diffusion embedding as the output response for output-informed diffusion maps, using Sensor 1 as the input manifold. In the resulting embedding, eigenvectors 1 and 2 capture system X, {\em while eigenvectors 3 and 8 capture system Y} (Figs. \ref{fig:output1}a and \ref{fig:output1}b).

We can also do the same thing using Sensor 2 as the input manifold and the results are presented in Appendix \ref{app:out_sensor2}.

In future work we will use the parametrization of the {\em uncommon} manifold of each sensor to construct level sets of said sensor that are consistent with an observation set of the other sensor.

\section{Summary and Outlook}

We have demonstrated  how we can find, in a data-driven way, common measurements between two (or, in principle, several) simultaneous measurement streams; our illustration was based on multiple observations (time series) from three nonlinear dynamical systems. This was accomplished through two alternative techniques: (a) Alternating Diffusion Maps and (b) the construction of Jointly Smooth Functions.
Importantly, after the correlated measurements across the two sensor streams were detected, we could learn (in several data-driven ways) a quantitative approximation of their relation. 

We also showed how this approach can give us {\em a sense of causality}, helping uncover a data-driven dynamic evolution model for the common features. 
This suggest our first possible avenue of further research: it will be interesting to consider that the two (or more) sets of measurements come from different scale observations of multiscale systems (e.g. atomistic scale and continuum scale simulations of the same system).  This should provide useful information regarding the appropriate level at which a useful closure should be attempted. 

We initially studied the ``clean channel'' case, where each measurement channel (pixel) comes either from the process of interest or (exclusively or!) the sensor-specific processes.
We then proceeded to the ``dirty channel'' case, where each channel (pixel) contains a function of the both process of interest {\em and} sensor specific information.
In this case -in principle- there is no identifiability across the observations: each set of measurements is consistent with an entire level set of measurements of the other. This provides a second possible direction of future research: given the probability distribution of the original data in their respective spaces, it should be possible --given a set of measurements from one of the observation processes- to construct not only the level set of consistent measurements of the other process, but also ``the right'' probability density {\em on} the corresponding consistent level set. 

In this work learning the transformation (in principle, a diffeomorphism) between corresponding measurements from the two (or more) observation processes was demonstrated --as proof of concept-- using data science/ML techniques that are broadly avalable and used in the case of relatively few (say two, three, four) channels/dimensions. A true challenge lies in detecting the existence of, and constructing, these transformations in high dimensions, e.g. through solving functional equations or Hamilton-Jacobi-Bellman equations in high dimensions \cite{hu2024tackling, darbon2016algorithms, azmi2021optimal, han2018solving, sirignano2018dgm, dolgov2021tensor}. The construction of modern computational techniques capable of this constitutes, by itself, an area of intense current research. 

\section{Acknowledgments}
The work of DWS and IGK was partially supported by the US DOE and the US AFOSR. FD was funded by the Deutsche Forschungsgemeinschaft (DFG, German Research Foundation) – project no. 468830823 and DFG-SPP-229 (associated). EDK was funded by the Luxembourg National Research Fund (FNR), grant reference 16758846. For the purpose of open access, the authors have applied a Creative Commons Attribution 4.0 International (CC BY 4.0) license to any Author Accepted Manuscript version arising from this submission.

\bibliographystyle{plain}
\bibliography{main}

\section{Appendix}

\section{Computational Methods}
\label{sec:computational methods}
This section briefly introduces the manifold learning technique ``Diffusion Maps,'' as well as a particular version of it, ``Alternating Diffusion Maps'' and the similar method of ``Jointly smooth functions.'' We also discuss a data-driven approach that helps decide whether a given data set can be described by a smooth input-output function: ``Local Linear Regression''.

\subsection{Manifold Learning: Diffusion Maps}\label{app:dmaps}
The goal of manifold learning is to discover underlying nonlinear structure in high-dimensional data. Diffusion maps\cite{Coifman2006a,Lafon2004} accomplishes this by constructing a discrete approximation of the Laplace-Beltrami operator on the data. When the data are sampled from a low-dimensional manifold, the discrete operator converges  (at the appropriate limit of infinite sample points) to the continuous Laplace-Beltrami operator on the manifold. The discrete operator is constructed by defining a weighted graph on the sampled data, where the weights between points i and j is given by
\begin{equation}
\label{dmaps}
w_{i,j}=exp\left(- \frac{d(\bm{y}_i,\bm{y}_j)^2}{\epsilon^2} \right),
\end{equation}
where $d( . , . )$ represents a chosen distance metric, and $\epsilon$ represents a distance scale below which samples are considered similar. A weight of 1 indicates that two samples are identical, while a weight close to 0 indicates that two samples are very dissimilar. After some normalization, the eigenvectors $\phi$ of the weight matrix provide a new coordinate system to describe the data. Distances in this coordinate system are referred to as diffusion distances. Eigenvectors which do not contribute to this distance (due to low eigenvalues) can be truncated, and the reduced set of eigenvectors can serve as a proxy for the intrinsic manifold coordinates.

\subsection{Alternating-Diffusion}

\begin{figure*}[!htp]
\centering

\begin{tabular}{cc}
\includegraphics[width=0.34\textwidth]{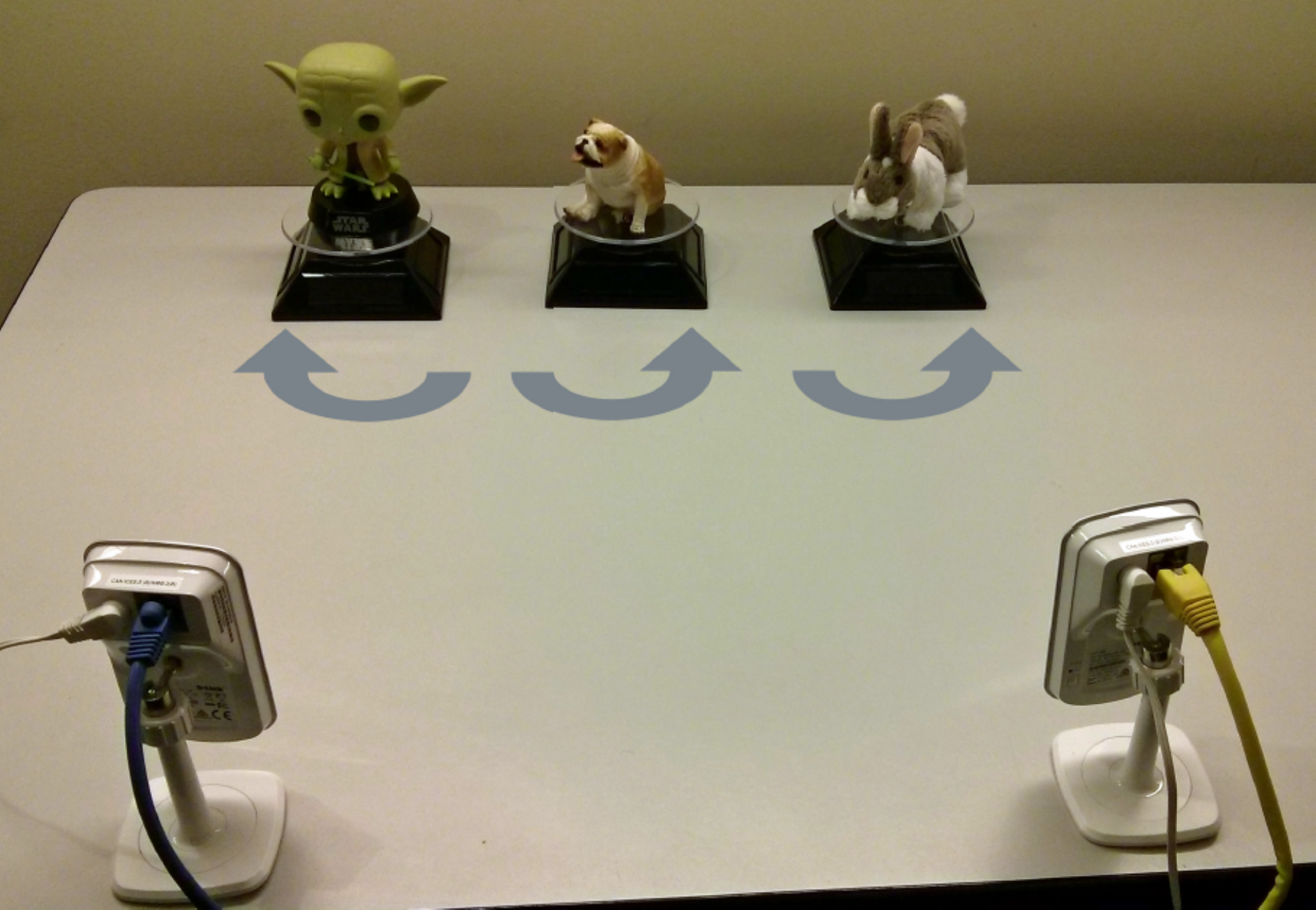} &
\includegraphics[width=0.65\textwidth]{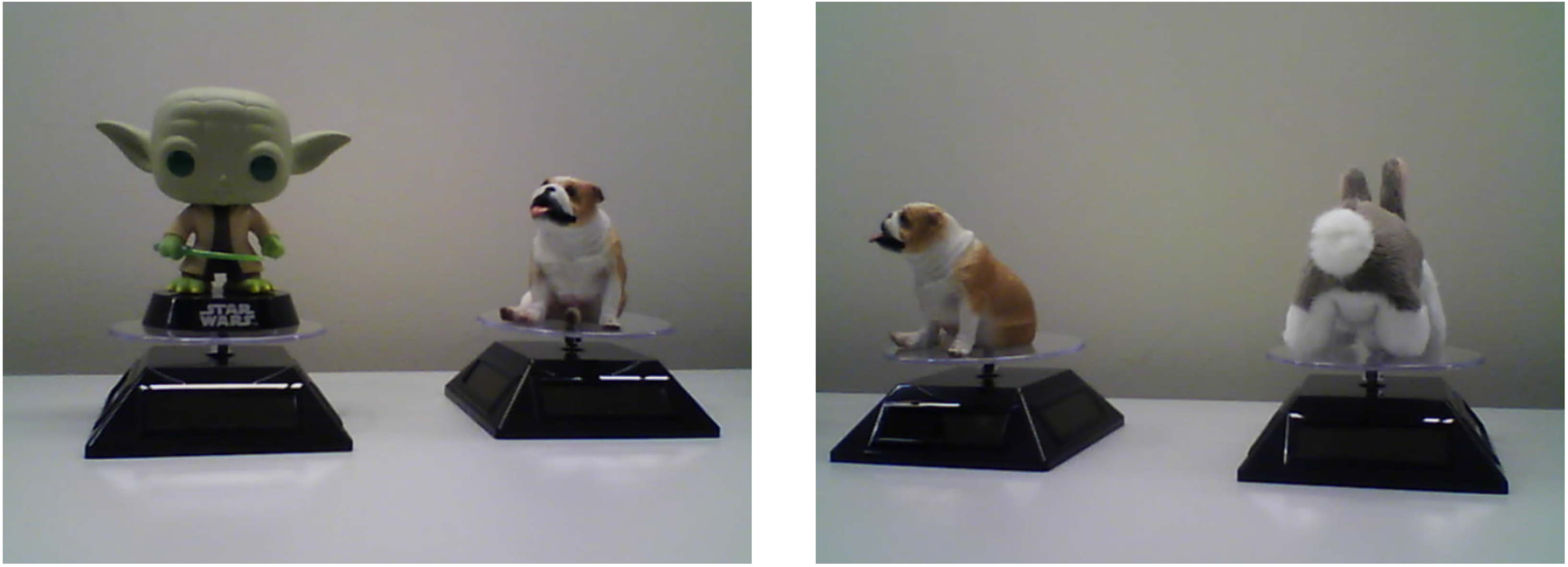} \\
(a) & (b)
\end{tabular}
\caption{The alternating diffusion algorithm is based upon data from two sensors. (a) A caricature setup where three figurines (Yoda, a bulldog, and a rabbit) are allowed to rotate independently and are photographed simultaneously by two cameras. (b) An example of simultaneous images from the two cameras. Although the bulldog is in the same intrinsic position in both images, the two cameras see different functions of its orientation. Alternating diffusion uncovers an embedding that is one-to-one with the intrinsic orientation.}
\label{fig:caricature}
\end{figure*}

The goal of alternating-diffusion\cite{lederman2018learning} is to handle the situation where two multi-dimensional sensors measure information about the same underlying system, but observations from each sensor are distorted by sensor-specific, uncorrelated noise. More precisely, suppose that we have three independent systems which can be described by the high-dimensional variables $X$, $Y$, and $Z$. We do not have access to these variables, but rather to a set of simultaneous measurements from two high-dimensional sensors $S^{(1)} = g(X,Y)$ and $S^{(2)} = h(X,Z)$. 
We require that $g$ and $h$ be bi-Lipschitz functions. The alternating-diffusion algorithm defines two weight matrices, one based on the measurements from $S^{(1)}$ and one based on the measurements from $S^{(2)}$, and constructs the alternating-diffusion operator as the product of the two normalized weight matrices. 
It has been shown that the diffusion process defined by this operator is equivalent to one that would have been created from measurements of only the common variable X. 
More details can be found in Ref. \cite{lederman2018learning}; we reproduce the procedure in Algorithm \ref{alg:altdmaps} and show a caricature example in Fig. \ref{fig:caricature}.
\begin{algorithm}[ht] 
    \hrulefill
    
    \textbf{\underline{Input}:} $2$ sets of $N$ simultaneous sensor measurements $\big\{ \vect{S}_{i}^{(1)},\vect{S}_{i}^{(2)}\big\} _{i=1}^{N}$
    where $\vect{S}_{i}^{(k)}\in\mathbb{R}^{d_{k}}$.
    
    \begin{enumerate}
        \item Calculate two affinity matrices:
        ${W}_{i,j}^{(1)}=\exp\left(-\frac{\big\Vert \vect{S}^{(1)}_{i}-\vect{S}^{(1)}_{j}\big\Vert^{2} }{2\sigma_{1}^{2}}\right)$,
        ${W}_{i,j}^{(2)}=\exp\left(-\frac{\big\Vert \vect{S}^{(2)}_{i}-\vect{S}^{(2)}_{j}\big\Vert^{2} }{2\sigma_{2}^{2}}\right)$
    
        \item Compute diffusion operators 

        $\tilde{W}_{i,j}^{(1)}=\frac{\displaystyle{W}_{i,j}^{(1)}}{\displaystyle\sum_{l=1} ^{N}{W}_{l,j}^{(1)}}$, $\tilde{W}_{i,j}^{(2)}=\frac{\displaystyle{W}_{i,j}^{(2)}}{\displaystyle\sum_{l=1} ^{N}{W}_{l,j}^{(2)}}$
        \item Compute the alternating diffusion operator $\tilde{W}=\tilde{W}^{(2)}\tilde{W}^{(1)}$
        \item Compute a low-dimensional embedding based on $\tilde{W}=\tilde{W}^{(2)}\tilde{W}^{(1)}$
        \end{enumerate}
    \caption{Alternating diffusion map embedding from simultaneous sensor measurements.}
    \label{alg:altdmaps}
    \hrulefill
\end{algorithm}

\subsection{Jointly Smooth Function Extraction} 
\label{sec:JSF}
We now introduce a different, recently developed, kernel based data driven approach to extract common directions in data sets: that of \emph{Jointly Smooth Functions} (JSFs)~\cite{Dietrich2022}. The JSF approach attempts to find functions of the individual sensor data sets that are \emph{jointly smooth} across \emph{all} the available data sets. We can then write all the common functions in terms of these JSFs, rather than describing the common parts of each data set as functions of each of the others.

Algorithm \ref{alg:jsf} constructs JSFs between $K$ data sets, arising from different observations of the same phenomenon, including sensor-specific (uncommon) noise. The key idea of the approximation procedure is to define function spaces on all $K$ data sets separately, through eigenfunctions of kernels. Then, we use singular value decomposition (SVD) to find the ``common'' functions across these spaces. For details on this approach, see the paper by Dietrich et al.~\cite{Dietrich2022}. Here, we have two data sets: $S^{(1)}$ and $S^{(2)}$. Therefore, we have to perform two eigendecompositions for two kernel matrices, and a subsequent SVD. The ``common'' functions between the two sensors $S^{(1)}$ and $S^{(2)}$ correspond to the common system (here, the limit cycle dynamics of system $Y$).

\begin{algorithm}[ht] 
    \hrulefill
    
    \textbf{\underline{Input}:} $K$ sets $\big\{ \vect{S}_{i}^{(1)},\vect{S}_{i}^{(2)},\dots\vect{S}_{i}^{(K)}\big\} _{i=1}^{N}$
    where $\vect{S}_{i}^{(k)}\in\mathbb{R}^{d_{k}}$.
    
    \textbf{\underline{Output}:} $M$ jointly smooth functions $\{ \vect{f}_{m}\in\mathbb{R}^{N}\} _{m=1}^{M}$.
    \begin{enumerate}
        \item For each observation set $\big\{ \vect{S}_{i}^{(k)}\big\} _{i=1}^{N}$
        compute the kernel:
        \[{K}_{k}(i,j)=\exp\left(-\frac{\big\Vert \vect{S}^{(k)}_{i}-\vect{S}^{(k)}_{j}\big\Vert^{2} }{2\sigma_{k}^{2}}\right)
        \]
        \item Compute $\textbf{W}_{k}\in\mathbb{R}^{N\times d}$, the first
        $d$ eigenvectors of $\textbf{K}_{k}$.
        \item Set $\textbf{W}=:\left[\textbf{W}_{1},\textbf{W}_{2},\dots,\textbf{W}_{K}\right]\in\mathbb{R}^{N\times Kd}$
        \item Compute the SVD decomposition: $\textbf{W}=\textbf{U} \textbf{$\Sigma$}\textbf{V}^{T}$
        \item Set $\vect{f}_{m}$ to be the $m$\textsuperscript{th} column of $\textbf{U}$.
    \end{enumerate}
    \caption{Jointly Smooth Functions from $K$ sets of observations.}
    \label{alg:jsf}
    \hrulefill
\end{algorithm}

\subsection{Local Linear Regression}

\begin{figure}
    \includegraphics[width=0.5\textwidth]{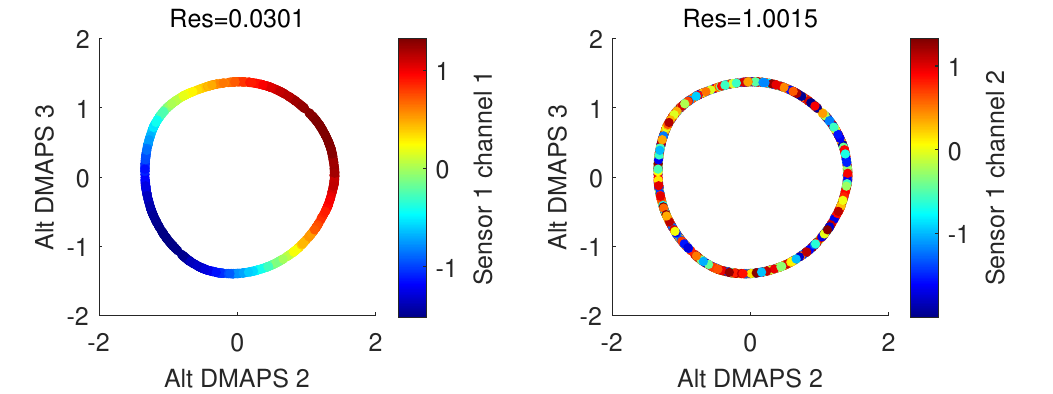}
    \centering
    \caption{\label{fig:example_LLR} Example plots of an alternating diffusion embedding colored by two different sensor coordinates, with the LLR residual above each plot. On the left, the coloring is smooth, and LLR gives a low residual, indicating that this coordinate belongs to the common system. On the right, the coloring is erratic, and the residual is about 1, indicating that this coordinate is influenced by an independent, non-common system.}
\end{figure}

When analyzing the eigenvectors of a diffusion operator, including the alternating-diffusion operator, simply discarding eigenvectors with eigenvalues lower than a defined threshold is not always sufficient to achieve the most parsimonious embedding. This is because higher harmonics of diffusion eigenfunctions are also eigenfunctions; e.g., $cos(x)$ and $cos(2x)$ are both eigenfunctions of the diffusion operator on a one-dimensional domain with no-flux boundary conditions. On multi-dimensional manifolds, the eigenvalues of these higher harmonics may happen to be higher than the eigenvalues corresponding to other unique coordinates.

To determine which eigenvectors of our discrete diffusion operator represent unique directions, we use local linear regression (LLR) as presented in Ref. \cite{dsilva2018parsimonious}. We attempt to fit each successive eigenvector as a locally linear function of the previous eigenvectors, where locality is defined by a Gaussian kernel. For each sample point $i$, we determine our local fit coefficients by minimizing the sum of squared errors, but weighting the squared error at each training point based on how similar that point is to our test point:
\begin{equation}
    \begin{gathered}
        \phi_k(i) \approx \alpha_k(i)+\bm{\beta}_k^T(i)\bm{\Phi}_{k-1}(i), \\
        \hat{\alpha}_k(i),\hat{\bm{\beta}}_k(i)= \\ \argmin_{\alpha,\bm{\beta}} \sum_{j\neq i}K(i,j){\left(\phi_k(j)-(\alpha+\bm{\beta} \bm{\Phi}_{k-1}(j))\right)}^2.
    \end{gathered}
\end{equation}
Eigenvectors with a low fit error are considered to represent higher harmonics of already known eigenvectors, and can be discarded, while eigenvectors with a high fit error represent new unique directions. We will also use this method to determine which of our original measurements can be fit as functions of our intrinsic manifold coordinates. In the alternating diffusion case, this means that those coordinates ``belong'' to the common system. We show an example of coordinates belonging to the common system, as well as coordinates not belonging to it in Fig. \ref{fig:example_LLR}.

\section{Data-driven approximation of functions}
\label{app:learning functions}

We now describe the approaches we used to learn functions between our identified, common coordinates and the original measurements.
\subsection{KNN}
\begin{figure}[!htp]

\includegraphics[width=0.35\textwidth]{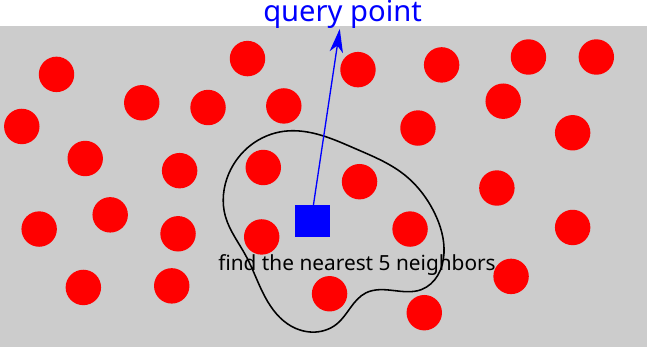} 
\centering
\caption{A schematic of the regression procedure with k-nearest neighbors. For any query point (blue square) in the test dataset, the five nearest neighbors (red circles in the loop) in the training set are identified. The label of the query point is then the weighted interpolation between the labels of neighbors. }
\label{fig:knn}
\end{figure}

Typical regression methods are based on some {\em a priori} assumptions on the topology of data, as well as, say, the degree of polynomials in curve fitting. 
A collection of methods, known as ``non-parametric regression'' methods exist, for which knowledge about the shape of data is not necessary.
The k-nearest neighbors (KNN~\cite{fix-1951}) method is a non-parametric technique, where the unknown label of a data point is estimated based on the labels of the $k$ nearest labeled points. In this section we first build a KDtree on the training data (with known labels). KDtree is a well-established algorithm for finding distances in high-dimensional data, where only similarity between close points are needed to be considered.
%
The location of a query (testing) point (with unknown label) is then identified in the constructed tree in $\theta_A(t)$ and $\theta_A(t-\Delta t)$ space (see Fig.~\ref{fig:knn}(a)).
The $k$ neighbor values of training $\theta_B(t)$ are then used to estimate the label of testing point. Here we used $k=5$ neighbours. In Fig.~\ref{fig:knn}(b) we show the true and prediction values for $\theta_B(t)$. The error is $\varepsilon_{kNN}=9.6\times 10^{-2}$.

\subsection{Geometric Harmonics}
Consider the case where we try to approximate a function $f(x)$ by another function $g_N(x)$, which takes the following form:
\begin{equation}
    f(x) \approx g_N(x) =\Sigma_{i=1}^N a_i \psi_i(x)
\end{equation}
Here, $g_N$ consists of a sum over $N$ orthogonal functions $\psi_i(x)$ weighted by some coefficients $a_i$. The well-known orthonormal basis function set in one dimension are the sine and cosine functions, which arise in the context of Fourier series. 

In previous sections, we presented diffusion maps (DMAPs) as a kernel learning method to find the intrinsic geometry of sets. Consider a (training) set $\Gamma$ subsampled from $\bar{\Gamma}$ with finite measure $\mu(\Gamma)<\infty$. The function $f:\Gamma\rightarrow\mathbb{R}$ is known and we are interested in approximating its value for some $x\notin \Gamma$ (this task is known as function extension/out-of-sample extension).
With no {\em a priori} assumption on the geometry of $\Gamma$, one can choose any class of functions for $\psi$. However, DMAPs can be used to set constraints on the feasibility of this extension based on the intrinsic geometry of the dataset.
Since the intrinsic geometry of $\Gamma$ is represented by DMAPs, the Nystr\"om method will allow us to extend $f$ outside the set $\Gamma$ using a special set of functions known as Geometric Harmonics \cite{Lafon2004, Coifman2006a}.
Consider that a function $f$ is represented by linear combination of kernels $k(x,y)$ in $\Gamma$, e.g.~$\int_\Gamma k(x,y) \phi_j(y)d\mu(y)=\lambda_j\phi_j$.
The basis ${\phi_j}$ of DMAP eigenfunctions, can now be extended to $\bar{\Gamma}$ using Nystr\"om:
\begin{equation}
    \forall \lambda_j>0, x\in \bar{\Gamma}, \psi_j(x):=\frac{1}{\lambda_j}\int_{\Gamma}k(x, y)\phi_j(y)d\mu(y)
\end{equation}
For $x\in \Gamma$ we have $\psi_j=\phi_j$, therefore ${\psi_j}$, Geometric Harmonics, are extensions of ${\phi_j}$ basis. Note, it can be shown that $\psi_j$ are orthonormal both in $\Gamma$ and $\bar{\Gamma}$, hence, can be used for the function approximation task above.
%
Here, our training dataset, $\Gamma$, is approximated by the top five eigenfunctions. $\theta_B(t)$ values for the test set, $\bar{\Gamma}$, are then approximated using a Geometric Harmonics code \cite{Bello-Rivas2017}. The error of the Geometric Harmonics regression based on \eqref{eq:error} is estimated at $\varepsilon= 1.3\times 10^{-2}$.
\subsection{Feed-Forward Neural Networks (FFNN)}
\begin{figure}[!htp]
\centering
\begin{tabular}{cc}
\includegraphics[width=0.2\textwidth]{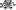} &
\includegraphics[width=0.2\textwidth]{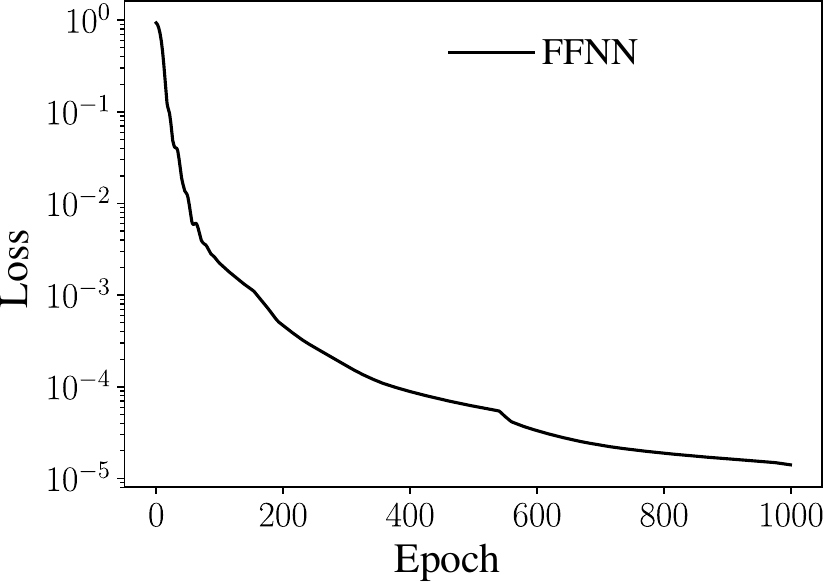}\\
(a) & (b)
\end{tabular}
\caption{(a) Schematic of Feed-Forward neural network with RReLU activation function. Inputs are training values of $\theta_A(t)$ and $\theta_A(t-\Delta t)$ and output is known values of $\theta_B(t)$. The network consists of two hidden layers with ten neurons in each layer. (b) The history of training in terms of MSE loss versus epoch.}
\label{fig:FFNN}
\end{figure}

In this section we use a multi-layered network of neurons to perform the regression task, i.e.~learning $\theta_B(t)$ as a function of $(\theta_A(t),\theta_A(t-\Delta t))$.
Our Feed-Forward Neural Network (FFNN) is shown schematically in Fig.~\ref{fig:FFNN}(a). Consisting of two hidden layers with ten neurons each, it is first initialized by random weights and then weights are corrected during each epoch by error backpropagation. We implement the network in PyTorch \cite{Paszke2017}, using the Adam optimizer for the correction of weights in each training epoch \cite{KingmaB14}.
The ``Randomized Leaky Rectified Liner Unit'' (RReLU) serves as our activation function; it has the following form, 
\begin{equation}
f_{act}(\alpha,x)=
\begin{cases}
\alpha x, & \text{for } x<0 \\ x, & \text{for } x \geq 0
\end{cases}
\end{equation}
In Fig.~\ref{fig:FFNN}(b) we show the history of training in terms of Mean Squares (MSE) loss as a function of epoch. After successful training we have used the network to find the values of $\theta_B(t)$ for the test dataset. The accuracy of prediction is shown in Fig.~\ref{fig:FFNN}(c), while the error based on \eqref{eq:error} is estimated at $\varepsilon=3.8\times10^{-4}$.

\section{Implementation of Jointly Smooth Functions}
\label{app:JSFres}

\begin{figure*} 
\centering
\begin{tabular}{cc}
\includegraphics[width=0.6\textwidth]{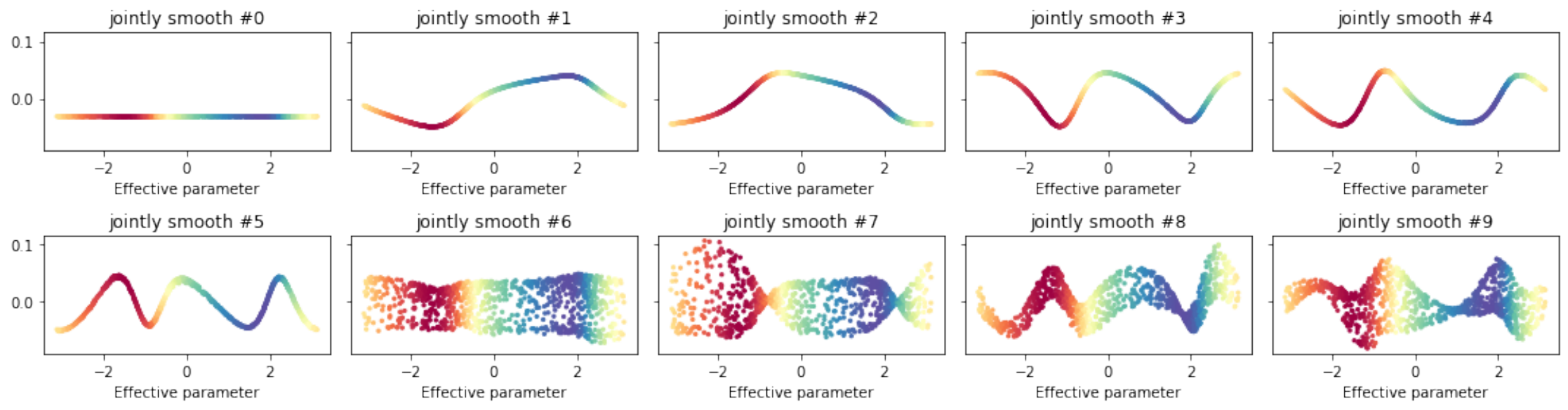} & 
\includegraphics[width=0.4\textwidth]{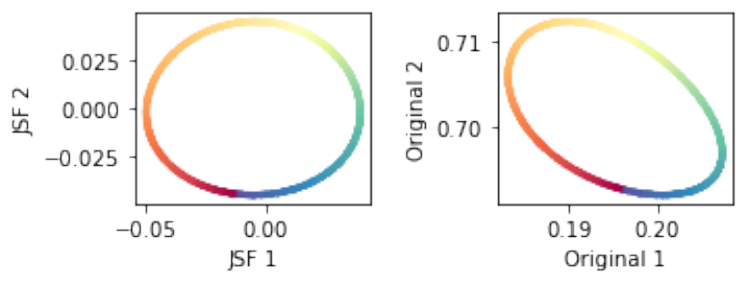} \\
    (a) & (b) 
\end{tabular}

\caption{(a) The first 10 extracted jointly smooth functions. (b)(Left)The embedding result for the two most parsimonious JSFs. \textit{(Right)} The original system X data colored by one JSF.}
\label{fig:jsf_caus1}
\end{figure*}

Here we present the results of the implementation of JSF to data from Setup 1 presented in Section \ref{sec:temporal}.
In Fig. \ref{fig:jsf_caus1}a, we visualize the first 10 jointly smooth functions. Similarly to Alternating Diffusion Maps, we can use LLR to select the two functions which give the most parsimonious embedding (see Fig. \ref{fig:jsf_caus1}b).

\begin{figure*} 
\centering
\begin{tabular}{cc}
\includegraphics[width=0.6\textwidth]{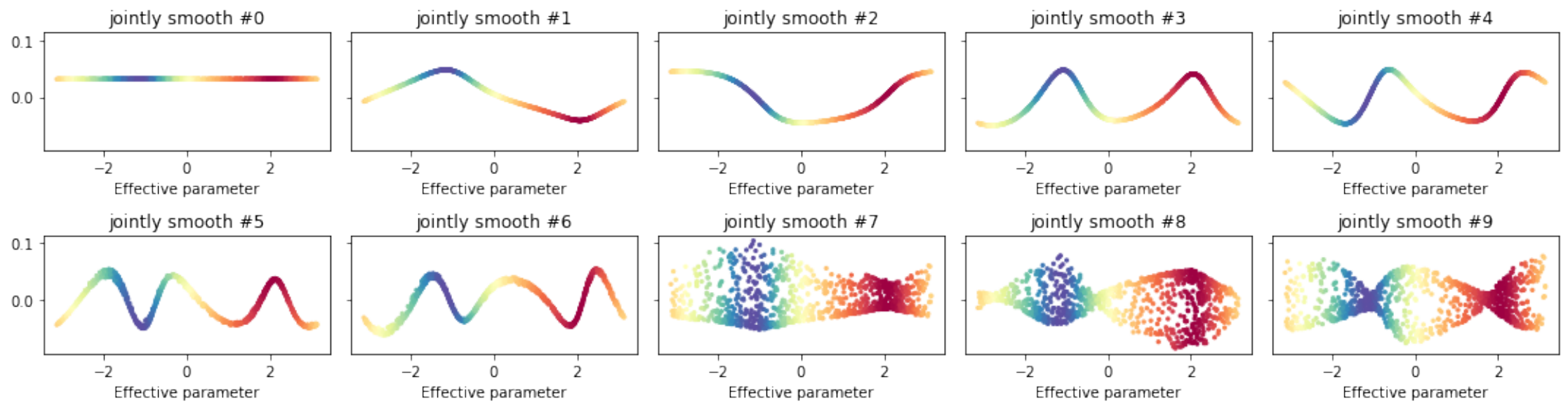} & 
\includegraphics[width=0.4\textwidth]{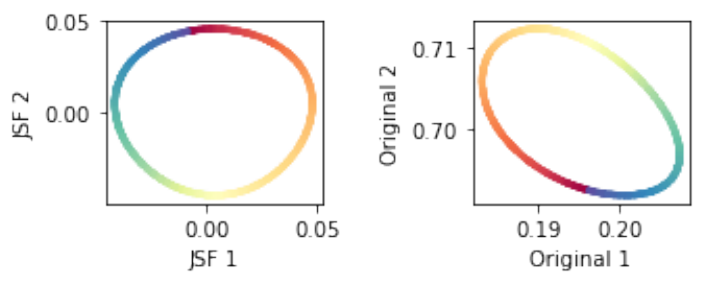} \\
    (a) & (b) 
\end{tabular}

\caption{(a) The first 10 extracted jointly smooth functions. (b)(Left)The embedding result for the two most parsimonious JSFs. \textit{(Right)} The original system X data colored by one JSF.}
\label{fig:jsf_caus2}
\end{figure*}

JSF is then implemented to data from Setup 2 presented in Section \ref{sec:temporal}. In Fig. \ref{fig:jsf_caus2}a, we visualize the first 10 jointly smooth functions. Similarly to Alternating Diffusion Maps, we can use LLR to select the two functions which give the most parsimonious embedding (see Fig. \ref{fig:jsf_caus2}b).

\begin{figure*} 
\centering
\begin{tabular}{cc}
\includegraphics[width=0.6\textwidth]{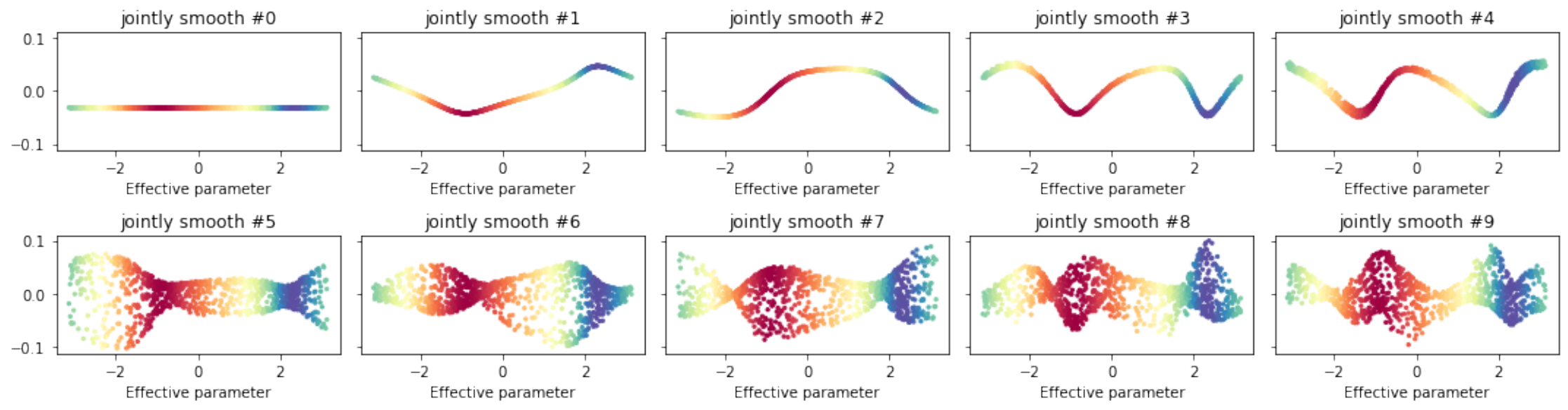} & 
\includegraphics[width=0.4\textwidth]{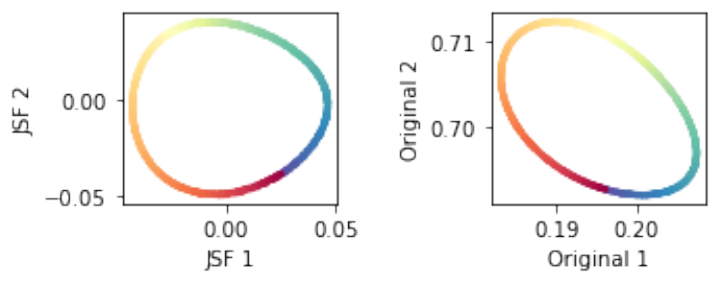} \\
    (a) & (b) 
\end{tabular}

\caption{(a) The first 10 extracted jointly smooth functions. (b)(Left)The embedding result for the two most parsimonious JSFs. \textit{(Right)} The original system X data colored by one JSF.}
\label{fig:jsf_mixed}
\end{figure*}

The results from the implementation of JSF to data described in Section \ref{sec:outputDMAPs} is presented here. In Fig.~\ref{fig:jsf_mixed}a, we visualize the first 10 jointly smooth functions. Similarly to Alternating Diffusion Maps, we can use LLR to select the two functions which give the most parsimonious embedding (see Fig.~\ref{fig:jsf_mixed}b).

\begin{figure*}
\centering
\begin{tabular}{cc}
    \includegraphics[width=0.35\textwidth]{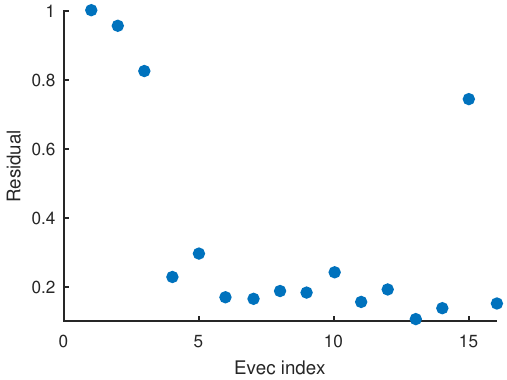} & 
\includegraphics[width=0.65\textwidth]{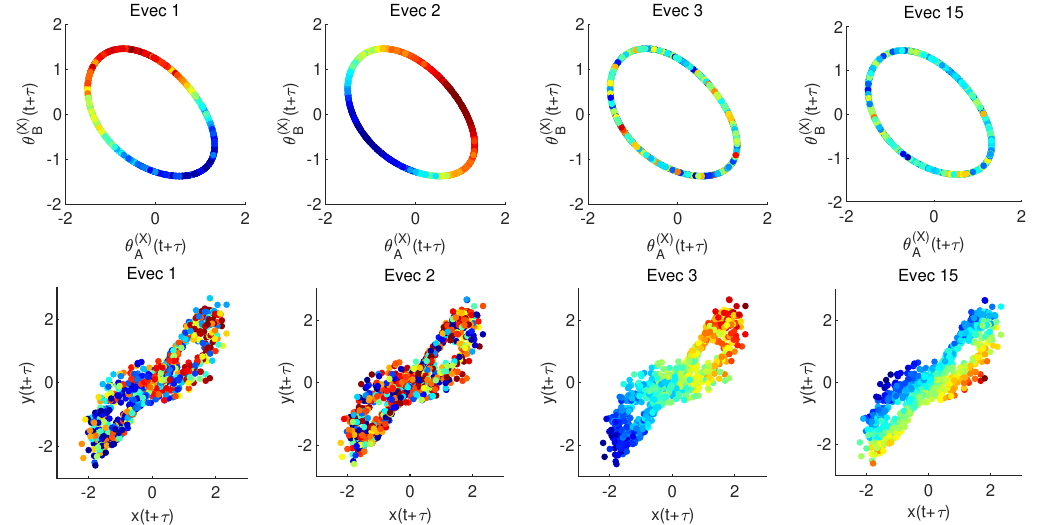} \\
    (a) & (b) 
\end{tabular}
\caption{(a) Results of running LLR on the set of successive eigenvectors $\phi_i$ (sorted by eigenvalue) from output diffusion maps on Sensor 2 data with the alternating-diffusion eigenvectors as the output. $\phi_1$ is trivially constant, and $\phi_2$ has a normalized LLR residual of 1 by definition. Eigenvectors 1, 2, 3, and 15 represent unique directions. (b)(Top row) Plots of the system X variables, colored by the output diffusion map eigenvectors 1, 2, 3, and 15. (Bottom row) Plots of the system Z variables, colored by the output diffusion map eigenvectors 1, 2, 3, and 15.}
\label{fig:output2}
\end{figure*}

\section{Output-informed DMAP results on Sensor 2}
\label{app:out_sensor2}
Similar to the implementation described in Section \ref{sec:outputDMAPs} for Sensor 1, we present the results for Sensor 2 data. In the resulting embedding, eigenvectors 1 and 2 capture system X, while eigenvectors 3 and 15 capture system Y (Figs. \ref{fig:output2}a and \ref{fig:output2}b).

\end{document}